\documentclass{article}

    \usepackage[final]{neurips_2021}

\usepackage[utf8]{inputenc} 
\usepackage[T1]{fontenc}    
\usepackage{hyperref}       
\usepackage{url}            
\usepackage{booktabs}       
\usepackage{amsfonts}       
\usepackage{nicefrac}       
\usepackage{microtype}      
\usepackage{xcolor}         

\usepackage{amsmath,amsfonts,amssymb,amsthm}
\usepackage{mathtools}
\usepackage{algorithm,algorithmic,caption}
\usepackage{enumitem}
\usepackage{textcomp}
\usepackage{booktabs}
\usepackage{multirow}
\usepackage{wrapfig}
\usepackage{stackrel}
\usepackage{etoc} 

\newcommand{\R}{\mathbb{R}}

\newcommand{\data}{\mathcal{D}}

\newcommand{\negSpace}{\vspace{-.05cm}}
\newcommand{\headSpace}{\vspace{-0.25cm}} 
\newcommand{\parSpace}{\vspace{-0.075cm}} 

\usepackage{xifthen}
\usepackage{amsmath,bm}
\newcommand{\normal}[3][]{%
    \ifthenelse{\isempty{#1}}
        {\mathcal{N}\left(#2, #3\right)}
        {\mathcal{N}\left(#1|#2, #3\right)}%
}

\newcommand{\norm}[1]{\left\lVert#1\right\rVert}

\usepackage[colorinlistoftodos]{todonotes}
\DeclareMathOperator{\diag}{diag}

\usepackage{mathtools}
\usepackage{cleveref}
\usepackage{textcomp}
\usepackage{booktabs}

\usepackage{caption}
\usepackage{subcaption}
\usepackage{tikz}
\usepackage{tcolorbox}

\makeatletter
\newcommand{\tikzmarkMath}[2]{%
    \tikz[%
        remember picture, 
        baseline = (#1.base),
        inner sep = 0pt,
        outer sep = 0pt, 
    ] \node (#1) {$\m@th\displaystyle #2$};%
}
\makeatother

\newtcbox{\mywboxtext}{on line,colback=white,colframe=black,size=fbox,arc=3pt,boxrule=0.8pt}
\newcommand{\mywboxmath}[1]{\mywboxtext{$#1$}}

\title{Distributional Gradient Matching for \\Learning Uncertain Neural Dynamics Models}

\author{%
Lenart Treven\thanks{Equal Contribution. Correspondence to \texttt{trevenl@ethz.ch}, \texttt{wenkph@ethz.ch}.}\\
ETH Z\"{u}rich\\
\texttt{trevenl@ethz.ch}\\
\And
Philippe Wenk\footnotemark[1]\\
ETH Z\"{u}rich\\
\texttt{wenkph@ethz.ch}\\
\And
Florian Dörfler\\
ETH Z\"{u}rich\\
\texttt{dorfler@ethz.ch}\\
\And
Andreas Krause\\
ETH Z\"{u}rich\\
\texttt{krausea@ethz.ch}\\
}

\begin{document}

\etocdepthtag.toc{mtchapter}
\etocsettagdepth{mtchapter}{subsection}
\etocsettagdepth{mtappendix}{none}

\maketitle

\begin{abstract}
    \looseness -1 Differential equations in general and neural ODEs in particular are an essential technique in continuous-time system identification. While many deterministic learning algorithms have been designed based on numerical integration via the adjoint method, many downstream tasks such as active learning, exploration in reinforcement learning, robust control, or filtering require accurate estimates of predictive uncertainties. In this work, we propose a novel approach towards estimating epistemically uncertain neural ODEs, avoiding the numerical integration bottleneck. Instead of modeling uncertainty in the ODE parameters, we directly model  uncertainties in the state space. Our algorithm -- {\em distributional gradient matching (DGM)} -- jointly trains a smoother and a dynamics model and matches their gradients via minimizing a Wasserstein loss. Our experiments show that, compared to traditional approximate inference methods based on numerical integration, our approach is faster to train, faster at predicting previously unseen trajectories, and in the context of neural ODEs, significantly more accurate.
\end{abstract}

\headSpace
\section{Introduction}
\headSpace
\label{sectinon: Introduction}
\begin{wrapfigure}{r}{0.5\textwidth}
  \begin{center}
  \vspace{-1.2cm}
     \includegraphics[width=0.5\textwidth]{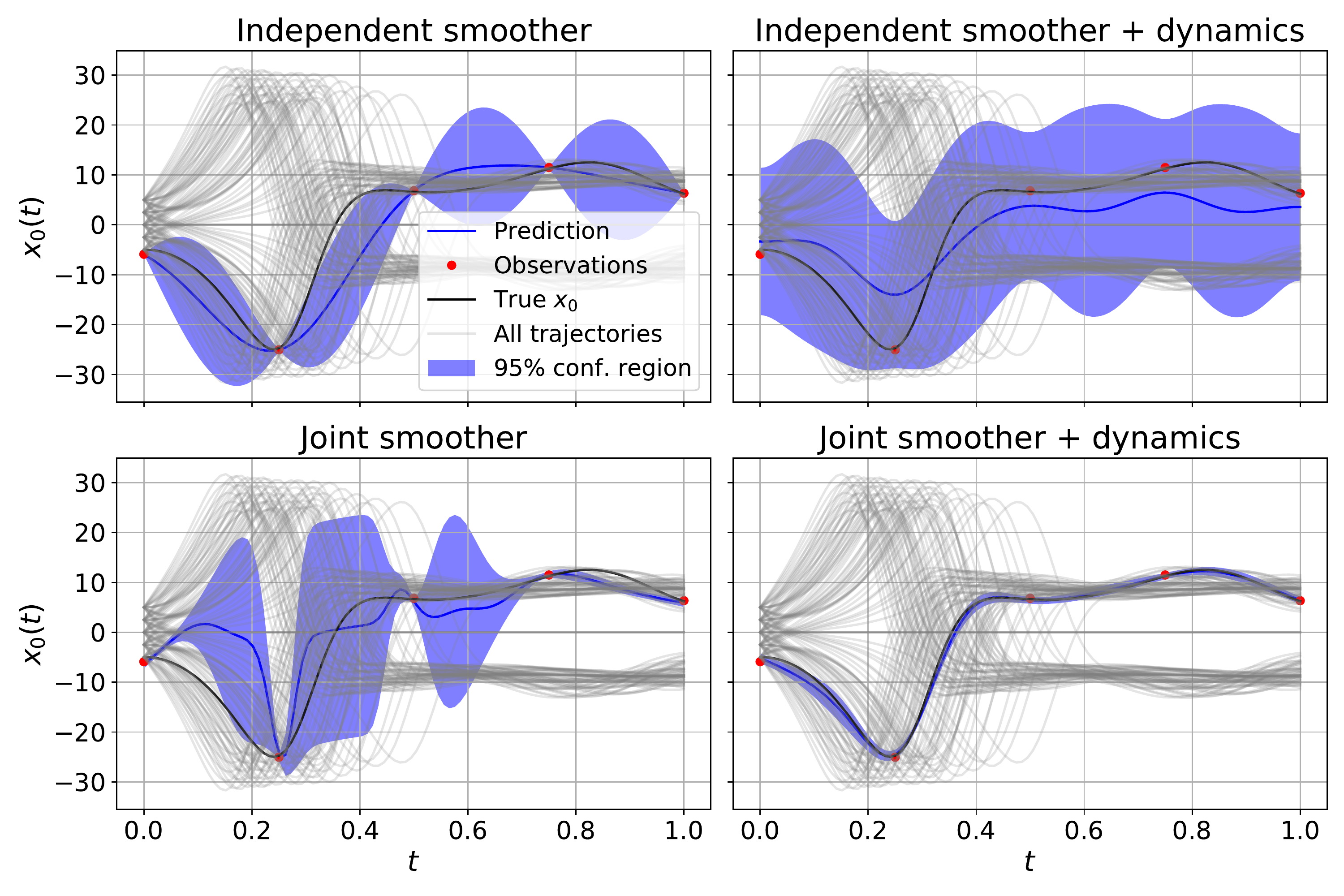}
  \end{center}
  \vspace{-0.4cm}
    \caption{\looseness -1 Illustration of DGM: Learning a joint smoother (first vs second row) across trajectories enables sharing observational data. Dynamics regularization (first vs second column) substantially improves prediction accuracy of joint smoother.}
    \label{figure: ablation study}
  \vspace{-0.4cm}
\end{wrapfigure}

For continuous-time system identification and control, ordinary differential equations form an essential class of models, deployed in applications ranging from robotics \citep{spong2006robot} to biology \citep{jones2009differential}. Here, it is assumed that the evolution of a system is described by the evolution of continuous state variables, whose time-derivative is given by a set of parametrized equations.
Often, these equations are derived from first principles, e.g., rigid body dynamics \citep{wittenburg2013dynamics}, mass action kinetics \citep{ingalls2013mathematical}, or Hamiltonian dynamics \citep{greydanus2019hamiltonian}, or chosen 
for computational convenience (e.g., linear systems \citep{Ljung1998}) or parametrized to facilitate system identification \citep{brunton2016discovering}.
\parSpace

\looseness-1
Such construction methods lead to intriguing properties, including guarantees on physical realizability \citep{wensing2017linear}, favorable convergence properties \citep{ortega2018dynamic}, or a structure suitable for downstream tasks such as control design \citep{ortega2002interconnection}. However, such models often capture the system dynamics only approximately, leading to a potentially significant discrepancy
between the model and reality \citep{ljung1999model}. 
Moreover, when expert knowledge is not available, or precise parameter values are cumbersome to obtain, system identification from raw time series data becomes necessary.
In this case, one may seek more expressive {\em nonparametric} models instead \citep{rackauckas2020universal,pillonetto2014kernel}. If the model is completely replaced by a neural network, the resulting model is called \emph{neural ODE} \citep{chen2018neural}.
Despite their large number of parameters, as demonstrated by \citet{chen2018neural,kidger2020hey,zhuang2020adaptive,zhuang2021mali}, deterministic neural ODEs can be efficiently trained, enabling accurate deterministic trajectory predictions. 
\begin{figure}
\begin{tcolorbox}
\begin{align*}
    &
    \tikzmarkMath{SmootherText}{
    \mywboxmath{
    \text{Smoother: }
    (t, x_0 )
    \xmapsto{GP_{\textcolor{red}{\varphi}}}
    {\textcolor{red}{p_{S}(x(t))}},
    {\textcolor{red}{p_{S}(\dot x(t))}} 
    }}
    \quad
    \tikzmarkMath{DynamicsText}{
    \mywboxmath{
    \text{Dynamics: } x(t)
    \xmapsto{NN_{\textcolor{blue}{\psi}}} \tikzmarkMath{Dynamics}{\textcolor{blue}{p_D(\dot x(t))}}}
    }
    \\[1em]
    & \quad \quad \quad \quad \quad \quad \quad \quad
     \mywboxmath{
    \tikzmarkMath{ObjectiveText}{
    \max_{\textcolor{blue}{\psi}, \textcolor{red}{\varphi}} \log \textcolor{red}{p_S(X_S)} - \lambda \cdot \mathbb{W}_2^2(\textcolor{red}{p_S(\dot X_S)}, \textcolor{blue}{p_D(\dot X_D)})
    }}
\end{align*}
\end{tcolorbox}

\begin{tikzpicture}[overlay, remember picture, shorten >=5pt, shorten <=0pt]
\draw[thick, ->, red](SmootherText.south) + (-2cm,0pt) to [out=-90,in=180] (ObjectiveText);
\draw[thick, ->, blue] (DynamicsText.south) + (1cm,0pt) to [out=-90,in=0] (ObjectiveText);

\end{tikzpicture}

\caption{High-level depiction of DGM.}
\label{fig: high-level overview}
\end{figure}
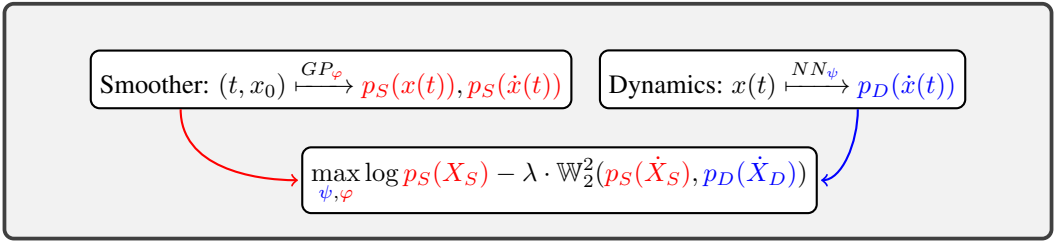
\parSpace
For many practical applications however, accurate \emph{uncertainty estimates} are essential, as they guide downstream tasks like reinforcement learning \citep{deisenroth2011pilco,schulman2015trust}, safety guarantees \citep{berkenkamp2017safe}, robust control design \citep{hjalmarsson2005experiment}, planning under uncertainty \citep{lavalle2006planning}, probabilistic forecasting in meteorology \citep{fanfarillo2021probabilistic}, or active learning / experimental design \citep{srinivas2009gaussian}. A common way of obtaining such uncertainties is via a Bayesian framework. However, as observed by \citet{dandekar2020bayesian}, Bayesian training of neural ODEs in a dynamics setting remains largely unexplored. They demonstrate that initial variational-based inference schemes for Bayesian neural ODEs suffer from several serious drawbacks and thus propose sampling-based alternatives. However, as surfaced by our experiments in \Cref{section: Experiments}, sampling-based approaches still exhibit serious challenges. These pertain both to robustness (even if highly informed priors are supplied), and reliance on frequent numerical integration of large neural networks, which poses severe computational challenges for many downstream tasks like sampling-based planning \citep{karaman2011sampling} or uncertainty propagation in prediction.
\parSpace

\paragraph{Contributions} In this work, we propose a novel approach for uncertainty quantification in nonlinear dynamical systems (cf.~\Cref{figure: ablation study}). Crucially, our approach avoids explicit costly and non-robust numerical integration, by employing a probabilistic smoother of the observational data, whose representation we learn jointly across multiple trajectories. To capture dynamics, we regularize our smoother with a dynamics model. Latter captures epistemic uncertainty in the gradients of the ODE, which we match with the smoother's gradients by minimizing a Wasserstein loss, hence we call our approach {\em Distributional Gradient Matching ($\operatorname{DGM}$)}.
In summary, our main contributions are: 
\negSpace \negSpace
\begin{itemize}
    \item We develop $\operatorname{DGM}$, an approach\footnote{Code is available at: \url{https://github.com/lenarttreven/dgm}} for capturing epistemic uncertainty about nonlinear dynamical systems by \emph{jointly} training a smoother and a neural dynamics model;
    \negSpace
    \item We provide a computationally efficient and statistically accurate mechanism for prediction, by focusing directly on the posterior / predictive state distribution.
    \negSpace
    \item We experimentally demonstrate the effectiveness of our approach on learning challenging, chaotic dynamical systems, and generalizing to new unseen inital conditions.
\end{itemize}

\paragraph{High-level overview} A high-level depiction of our algorithm is shown in Figure \ref{fig: high-level overview}. In principle, $\operatorname{DGM}$ jointly learns a smoother (S) and a dynamics model (D). The smoother model, chosen to be a Gaussian process, maps an initial condition ${x}_0$ and a time $t$ to the state distribution $p_S({x}(t))$ and state derivatives distribution $p_S(\dot{{x}}(t))$ reached at that time. The dynamics model, chosen to be a neural network, represents an ODE that maps states ${x}(t)$ to the derivative distribution $p_D(\dot{{x}}(t))$. Both models are evaluated at some training times and all its output distributions collected in the random variables $X_S$, $\dot{X}_S$ and $\dot{X}_D$. The parameters of these models are then jointly trained using a Wasserstein-distance-based objective directly on the level of distributions. For more details on every one of these components, we refer to \Cref{section: Distributional GM Key Ideas}. There, we introduce all components individually and then present how they interplay. \Cref{section: Distributional GM Key Ideas} builds on known concepts from the literature, which we summarize in \Cref{section: Background}. Finally, in \Cref{section: Experiments}, we present the empirical study of the DGM algorithm, where we benchmark it against the state-of-the-art, uncertainty aware dynamics models.

\parSpace

\headSpace
\section{Background}
\headSpace
\label{section: Background}
\subsection{Problem Statement}
\parSpace
Consider a continuous-time dynamical system whose $K$-dimensional state $\bm{x} \in \R^K$ evolves according to an unknown ordinary differential equation of the form
\negSpace
\begin{equation}
    \label{equation: Differential Equation Ground Truth}
    \dot{\bm{x}} = \bm{f}^*(\bm{x}).
\end{equation}
\negSpace
Here, $\bm{f}^*$ is an arbitrary, unknown function assumed to be locally Lipschitz continuous, to guarantee existence and uniqueness of trajectories for every initial condition. In our experiment, we initialize the system at $M$ different initial conditions $\bm{x}_m(0)$, $m \in \{1,\dots, M\}$,  and let it evolve to generate $M$ trajectories. Each trajectory is then observed at discrete (but not necessarily uniformly spaced) time-points, where the number of observations $\left(N_{m} \right)_{m \in\{1\dots M\}}$ can vary from trajectory to trajectory. Thus, a trajectory $m$ is described by its initial condition $\bm{x}_m(0)$, and the observations $\bm{y}_m \coloneqq \left[\bm{x}_m(t_{n, m}) + \bm{\epsilon}_{n, m}\right]_{n \in\{1 \dots N_m\}}$ at times $\bm{t}_m \coloneqq \left[t_{n, m}\right]_{n \in\{1\dots N_m\}}$, where the additive observation noise $\bm{\epsilon}_{n, m}$ is assumed to be drawn i.i.d. from a zero mean Gaussian, whose covariance is given by $\bm{\Sigma}_{\bm{\epsilon}} \coloneqq \text{diag}(\sigma_1^2, \dots, \sigma_K^2)$. We denote by $\data$ the dataset, consisting of $M$ initial conditions $\bm{x}_m(0)$, observation times $\bm{t}_m$, and observations $\bm{y}_m$.
\parSpace

To model the unknown dynamical system, we choose a parametric Ansatz $\dot{\bm{x}} = \bm{f}(\bm{x}, \bm{\theta})$. Depending on the amount of expert knowledge, this parameterization can follow a white-box, gray-box, or black-box methodology \citep{bohlin2006practical}. In any case, the parametric form of $\bm{f}$ is fixed a priori (e.g., a neural network), and the key challenge is to infer a reasonable distribution over the parameters $\bm{\theta}$, conditioned on the data $\data$. For later tasks, we are particularly interested in the \emph{predictive posterior state distribution} 
\negSpace
\begin{equation}
\label{equation: predictive posterior distribution}
    p(\bm{x}_{\text{new}}(\bm{t}_{\text{new}}) | \data, \bm{t}_{\text{new}}, \bm{x}_{\text{new}}(0)),
\end{equation} 
i.e., the posterior distribution of the states starting from a potentially unseen initial condition $\bm{x}_{\text{new}}(0)$ and evaluated at times $\bm{t}_{\text{new}}$. This posterior would then be used by the downstream or prediction tasks described in the introduction.
\parSpace

\subsection{Bayesian Parameter Inference}
\parSpace
In the case of Bayesian parameter inference, an additional prior $p(\bm{\theta})$ is imposed on the parameters $\bm{\theta}$ so that the posterior distribution of \Cref{equation: predictive posterior distribution} can be inferred.
Unfortunately, this distribution is not analytically tractable for most choices of $\bm{f}(\bm{x}, \bm{\theta})$, which is especially true when we model $\bm{f}$ with a neural network. Formally, for fixed parameters $\bm{\theta}$, initial condition $\bm{x}(0)$ and observation time $t$, the likelihood of an observation $\bm{y}$ is given by
\begin{equation}
\label{equation: Bayesian parametric model}
    p(\bm{y}(t) | \bm{x}(0), t, \bm{\theta}, \bm{\Sigma_{\text{obs}}}) = \mathcal{N}\left(\bm{y}(t) \middle | \bm{x}(0) + \int_0^t \bm{f}(\bm{x}(\tau), \bm{\theta}) d\tau, \bm{\Sigma_{\text{obs}}}\right).
\end{equation}
Using the fact that all noise realizations are independent, the expression \eqref{equation: Bayesian parametric model} can be used to calculate the likelihood of all observations in $\data$. Most state-of-the-art parameter inference schemes use this fact to create samples $\hat{\bm{\theta}}_s$ of the posterior over parameters $p(\bm{\theta}|\data)$ using various Monte Carlo methods. Given a new initial condition $\bm{x}(0)$ and observation time $t$, these samples $\hat{\bm{\theta}}_s$ can then be turned into samples of the predictive posterior state again by numerically integrating
\begin{equation}
\label{equation: MC Prediction}
    \hat{\bm{x}}_s(t) = \bm{x}(0) + \int_0^t \bm{f}(\bm{x}(\tau), \hat{\bm{\theta}}_s) d\tau.
\end{equation}
Clearly, both training (i.e., obtaining the samples $\hat{\bm{\theta}}_s$) and prediction (i.e., evaluating \Cref{equation: MC Prediction}) require integrating the system dynamics $\bm{f}$ many times. Especially when we model $\bm{f}$ with a neural network, this can be a huge burden, both numerically and computationally \citep{kelly2020learning}.
\parSpace

\looseness -1 As an alternative approach, we can approximate the posterior $p(\bm{\theta} | \data)$ with variational inference \citep{bishop2006pattern}. However, we run into similar bottlenecks. While optimizing the variational objective, e.g., the ELBO, many integration steps are necessary to evaluate the unnormalized posterior. Also, at inference time, to obtain a distribution over state $\hat{\bm{x}}_s(t)$, we still need to integrate $\bm{f}$ several times. Furthermore, \citet{dandekar2020bayesian}
report poor forecasting performance by the variational approach.
\parSpace

\headSpace
\section{Distributional Gradient Matching}
\headSpace
\label{section: Distributional GM Key Ideas}
In both the Monte Carlo sampling-based and variational approaches, all information about the dynamical system is stored in the estimates of the system parameters $\hat{\bm{\theta}}$. This makes these approaches rather cumbersome: Both for obtaining estimates of $\hat{\bm{\theta}}$ and for obtaining the predictive posterior over states, once $\hat{\bm{\theta}}$ is found, we need multiple rounds of numerically integrating a potentially complicated (neural) differential equation. We thus have identified two bottlenecks limiting the performance and applicability of these algorithms: namely, numerical integration of $\bm{f}$ and inference of the system parameters $\bm{\theta}$. In our proposed algorithm, we {\em avoid both} of these bottlenecks by directly working with the posterior distribution in the state space. 
\parSpace

To this end, we introduce a probabilistic, differentiable \emph{smoother model}, that directly maps a tuple $(t, \bm{x}(0))$ consisting of a time point $t$ and an initial condition $\bm{x}(0))$ as input and maps it to the corresponding distribution over $\bm{x}(t)$. Thus, the smoother directly replaces the costly, numerical integration steps, needed, e.g., to evaluate \Cref{equation: predictive posterior distribution}.
\parSpace

Albeit computationally attractive, this approach has one serious drawback. Since the smoother no longer explicitly integrates differential equations, there is no guarantee that the obtained smoother model follows any vector field. Thus, the smoother model is strictly more general than the systems described by \Cref{equation: Differential Equation Ground Truth}. Unlike ODEs, it is able to capture mappings whose underlying functions violate, e.g., Lipschitz or Markovianity properties, which is clearly not desirable. To address this issue, we introduce a regularization term, $\mathcal{L}_{\text{dynamics}}$, which ensures that a trajectory predicted by the smoother is encouraged to follow some underlying system of the form of \Cref{equation: Differential Equation Ground Truth}. The smoother is then trained with the multi-objective loss function 
\begin{equation}
\label{equation: Simplified Loss}
    \mathcal{L} \coloneqq \mathcal{L}_{\text{data}} + \lambda \cdot \mathcal{L}_{\text{dynamics}},
\end{equation}
where, $\mathcal{L}_{\text{data}}$ is a smoother-dependent loss function that ensures a sufficiently accurate data fit, and $\lambda$ is a trade-off parameter.
\parSpace

\subsection{Regularization by Matching Distributions over Gradients}
\parSpace
\looseness-1
To ultimately define $\mathcal{L}_{\text{dynamics}}$, first choose a parametric \emph{dynamics model} similar to $\bm{f}(\bm{x}, \bm{\theta})$ in \Cref{equation: Bayesian parametric model}, that maps states to their derivatives. Second, define a set of \emph{supporting points} $\mathcal{T}$ with the corresponding \emph{supporting gradients} $\dot{\mathcal{X}}$ as
\vspace{-0.05cm}
\begin{align*}
 \mathcal{T} \coloneqq
    \left\{
    \left(t_{\text{supp},l},
    \bm{x}_{\text{supp},l}(0)
    \right)_{l \in \{1 \dots N_{\text{supp}}\}}
    \right\} ,\quad
    \dot{\mathcal{X}} \coloneqq \left\{
    \left(
    \dot{\bm{x}}_{\text{supp},l}
    \right)_{l \in \{1 \dots N_{\text{supp}}\}}
    \right\}.
\end{align*}
\negSpace
Here, the $l$-th element represents the event that the dynamical system's derivative at time $t_{\text{supp},l}$ is $\dot{\bm{x}}_{\text{supp},l}$, after being initialized at time $0$ at initial condition $\bm{x}_{\text{supp},l}(0)$.
\parSpace

\looseness -1 Given both the smoother and the dynamics model, we have now two different ways to calculate distributions over $\dot{\mathcal{X}}$ given some data $\mathcal{D}$ and supporting points $\mathcal{T}$. First, we can directly leverage the differentiability and global nature of our smoother model to extract a distribution $p_S(\dot{\mathcal{X}}| \mathcal{D}, \mathcal{T})$ from the smoother model. Second, we can first use the smoother to obtain state estimates and then plug these state estimates into the dynamics model, to obtain a second distribution $p_D(\dot{\mathcal{X}} | \mathcal{D}, \mathcal{T})$. Clearly, if the solution proposed by the smoother follows the dynamics, these two distributions should match. Thus, we can regularize the smoother to follow the solution of \Cref{equation: Bayesian parametric model} by defining $\mathcal{L}_{\text{dynamics}}$ to encode the \emph{distance} between $p_D(\dot{\mathcal{X}} | \mathcal{D}, \mathcal{T})$ and $p_S(\dot{\mathcal{X}}| \mathcal{D}, \mathcal{T})$ to be small in some metric. By minimizing the overall loss, we thus match the distributions over the gradients of the smoother and the dynamics model.
\parSpace

\subsection{Smoothing jointly over Trajectories with Deep Gaussian Processes }
\parSpace
\looseness -1  The core of $\operatorname{DGM}$ is formed by a smoother model. In principle, the posterior state distribution of \Cref{equation: predictive posterior distribution} could be modeled by any Bayesian regression technique. However, calculating $p_{S}(\dot{\mathcal{X}} | \mathcal{D}, \mathcal{T})$ is generally more involved. Here, the key challenge is evaluating this posterior, which is already computationally challenging, e.g., for simple Bayesian neural networks. For Gaussian processes, however, this becomes straightforward, since derivatives of GPs remain GPs \citep{solak2003derivative}. Thus, $\operatorname{DGM}$ uses a GP smoother. For scalability and simplicity, we keep $K$ different, independent smoothers, one for each state dimension. However, if computational complexity is not a concern, our approach generalizes directly to multi-output Gaussian processes. Below, we focus on the one-dimensional case, for clarity of exposition.
For notational compactness, all vectors with a superscript should be interpreted as vectors over time in this subsection. For example, the vector $\bm{x}^{(k)}$ consists of all the $k$-th elements of the state vectors $\bm{x}(t_{n, m}), n \in \{1, \dots, N_m\}, m \in \{1, \dots, M\}$.
\parSpace

 We define a Gaussian process with a differentiable mean function $\mu(\bm{x}_m(0), t_{n,m})$ as well as a differentiable and positive-definite kernel function $\mathcal{K}_{\text{RBF}}(\bm{\phi}(\bm{x}_m(0), t_{n, m}), \bm{\phi}(\bm{x}_{m'}(0), t_{n', m'})$. Here, the kernel is given by the composition of a standard ARD-RBF kernel \citep{Rasmussen2004} and a differentiable feature extractor $\bm{\phi}$ parametrized by a deep neural network, as introduced by \citet{wilson2016deep}.
Following \citet{solak2003derivative}, given fixed $\bm{x}_{\text{supp}}$, 
we can now calculate the joint density of $(\dot{\bm{x}}^{(k)}_{\text{supp}}, \bm{y}^{(k)})$ for each state dimension $k$. Concatenating vectors accordingly across time and trajectories, let
\negSpace
\begin{align*}
\bm{\mu}^{(k)} &\coloneqq \mu^{(k)}\left(\bm{x}(0), \bm{t}\right), \quad
\dot{\bm{\mu}}^{(k)} \coloneqq \frac{\partial}{\partial t} \mu^{(k)}\left(\bm{x}_{\text{supp}}(0), \bm{t}_{\text{supp}}\right), \\
\bm{z}^{(k)} &\coloneqq \phi^{(k)}(\bm{x}(0), \bm{t}), \quad \,\,\,\,
\bm{z}^{(k)}_{\text{supp}} \coloneqq \phi^{(k)}(\bm{x}_{\text{supp}}(0), \bm{t}_{\text{supp}}), \\
\bm{\mathcal{K}}^{(k)} \coloneqq \mathcal{K}_{\text{RBF}}^{(k)}(\bm{z}^{(k)}, \bm{z}^{(k)}),& \quad
\dot{\bm{\mathcal{K}}}^{(k)} \coloneqq \frac{\partial}{\partial t_1}\mathcal{K}_{\text{RBF}}^{(k)}(\bm{z}^{(k)}_{\text{supp}}, \bm{z}^{(k)}), \quad
\ddot{\bm{\mathcal{K}}}^{(k)} \coloneqq \frac{\partial^2}{\partial t_1 \partial t_2} \mathcal{K}_{\text{RBF}}^{(k)}(\bm{z}^{(k)}_{\text{supp}}, \bm{z}^{(k)}_{\text{supp}}).
\end{align*}
\negSpace
Then the joint density of $(\dot{\bm{x}}^{(k)}_{\text{supp}}, \bm{y}^{(k)})$ can be written as
\negSpace
\begin{equation}
    \begin{pmatrix}
    \dot{\bm{x}}^{(k)}_{\text{supp}} \\
    \bm{y}^{(k)}
    \end{pmatrix}
    \sim 
    \normal{
    \begin{pmatrix}
    \dot{\bm{\mu}}^{(k)} \\
    \bm{\mu}^{(k)}
    \end{pmatrix}
    }{
    \begin{pmatrix}
    \ddot{\bm{\mathcal{K}}}^{(k)} & \dot{\bm{\mathcal{K}}}^{(k)} \\
    (\dot{\bm{\mathcal{K}}}^{(k)})^\top & \bm{\mathcal{K}}^{(k)}+ \sigma_k^2 \bm{I}
    \end{pmatrix}
    }.
\end{equation}
\parSpace

Here we denote by $\frac{\partial}{\partial t_1}$ the partial derivative with respect to time in the first coordinate, by $\frac{\partial}{\partial t_2}$ the partial derivative with respect to time in the second coordinate, and with $\sigma_k^2$ the corresponding noise variance of $\bm{\Sigma}_{\text{obs}}$.
\parSpace

Since the conditionals of a joint Gaussian random variable are again Gaussian distributed, $p_S$ is again Gaussian, i.e., $ p_S(\dot{\mathcal{X}}_k | \mathcal{D}, \mathcal{T}) = \normal[\dot{\bm{x}}^{(k)}_{\text{supp}}]{\bm{\mu}_S}{\bm{\Sigma}_S}$ with
\begin{equation}
    \begin{aligned}
    \bm{\mu}_S &\coloneqq \dot{\bm{{\mu}}}^{(k)} +  \dot{\bm{\mathcal{K}}}^{(k)}(\bm{\mathcal{K}}^{(k)} + \sigma_k^2 \bm{I})^{-1}\left(\bm{y}^{(k)} - \bm{\mu}^{(k)}\right), \\
    \bm{\Sigma}_S &\coloneqq \ddot{\bm{\mathcal{K}}}^{(k)} - \dot{\bm{\mathcal{K}}}^{(k)}(\bm{\mathcal{K}}^{(k)} + \sigma_k^2\bm I)^{-1}(\dot{\bm{\mathcal{K}}}^{(k)})^\top.
    \end{aligned} 
\end{equation}
Here, the index $k$ is used to highlight that this is just the distribution for one state dimension. To obtain the final $p_S(\dot{\mathcal{X}} | \mathcal{D}, \mathcal{T})$, we take the product over all state dimensions $k$.
\parSpace

To fit our model to the data, we minimize the negative marginal log likelihood of our observations, neglecting purely additive terms \citep{Rasmussen2004}, i.e.,
\begin{equation}
    \mathcal{L}_{\text{data}}
    \coloneqq
    \sum_{k=1}^{K}
    \frac{1}{2} \left(\bm{y}^{(k)} - \bm{\mu}^{(k)} \right)^\top
    \left(\bm{\mathcal{K}}^{(k)} + \sigma_k^2 \bm{I}\right)^{-1}
    \left(\bm{y}^{(k)} - \bm{\mu}^{(k)} \right)
    +\frac{1}{2} \operatorname{logdet}\left(\bm{\mathcal{K}}^{(k)} + \sigma_k^2 \bm{I}\right).
\end{equation}
\parSpace
Furthermore, the predictive posterior for a new point $x^{(k)}_{\text{test}}$ given time $t_{\text{test}}$ and initial condition $x^{(k)}_{\text{test}}(0)$ has the closed form
\begin{align}
    p_{S}(x^{(k)}_{\text{test}} | \mathcal{D}_k, t_{\text{test}}, \bm{x}_{\text{test}})
    =
    \mathcal{N}\left( x^{(k)}_{\text{test}}
    \middle | \mu^{(k)}_{\text{post}},
    \sigma^2_{\text{post}, k}
    \right),
\end{align}
\vspace{-0.3cm}
\begin{align}
    \text{where} \qquad \mu^{(k)}_{\text{post}} &= 
    \mu^{(k)}(\bm{x}_{\text{test}}(0), t_{\text{test}}) + \mathcal{K}^{(k)}_{\text{RBF}}(\bm{z}^{(k)}_{\text{test}}, \bm{z}^{(k)})^\top (\bm{\mathcal{K}}^{(k)} + \sigma_k^2 \bm{I})^{-1}
\left(\bm{y}^{(k)} - \bm{\mu}^{(k)}\right), \\
    \sigma^2_{\text{post}, k} &= \mathcal{K}^{(k)}_{\text{RBF}}(\bm{z}_{\text{test}}, \bm{z}_{\text{test}}) - \mathcal{K}^{(k)}_{\text{RBF}}(\bm{z}^{(k)}_{\text{test}}, \bm{z}^{(k)})^\top
    (\bm{\mathcal{K}}^{(k)} + \sigma_k^2 \bm{I})^{-1}\mathcal{K}^{(k)}_{\text{RBF}}
    (\bm{z}^{(k)}_{\text{test}}, \bm{z}^{(k)}).
\end{align}
\parSpace
\subsection{Representing Uncertainty in the Dynamics Model via the Reparametrization Trick}
\parSpace
As described at the beginning of this section, a key bottleneck of standard Bayesian approaches is the potentially high dimensionality of the dynamics parameter vector $\bm{\theta}$. The same is true for our approach. If we were to keep track of the distributions over all parameters of our dynamics model, calculating $p_{D}(\dot{\mathcal{X}} | \mathcal{D}, \mathcal{T})$ quickly becomes infeasible. 
\parSpace

However, especially in the case of modeling $\bm{f}$ with a neural network, the benefits of keeping distributions directly over $\bm{\theta}$ is unclear due to overparametrization. For both the downstream tasks and our training method, we are mainly interested in the distributions in the state space. Usually, the state space is significantly lower dimensional compared to the parameter space of $\bm\theta$. Furthermore, since the exact posterior state distributions are generally intractable, they normally have to be approximated anyways with simpler distributions for downstream tasks \citep{schulman2015trust, houthooft2016vime, berkenkamp2017safe}.
Thus, we change the parametrization of our dynamics model as follows. Instead of working directly with $\dot{\bm{x}}(t) = \bm{f}(\bm{x}(t), \bm{\theta})$ and keeping a distribution over $\bm{\theta}$, we model uncertainty directly on the level of the vector field as
\begin{equation}
    \label{equation: dynamics model}
    \dot{\bm{x}}(t) = \bm{f}(\bm{x}(t), \bm{\psi}) + \bm{\Sigma}_D^{\frac{1}{2}}(\bm{x}(t), \bm{\psi})\bm{\epsilon},
\end{equation}
where $\bm\epsilon \sim \mathcal{N}(0, \bm{I}_K)$ is drawn once per rollout (i.e., fixed within a trajectory) and $\bm{\Sigma}_D$ is a state-dependent and positive semi-definite matrix parametrized by a neural network. Here, $\bm{\psi}$ are the parameters of the new dynamics model, consisting of both the original parameters $\bm{\theta}$ and the weights of the neural network parametrizing $\bm{\Sigma}_D$. To keep the number of parameters reasonable, we employ a weight sharing scheme, detailed in \Cref{section: implementation details of DGM}.
\parSpace

In spirit, this modeling paradigm is very closely related to standard Bayesian training of NODEs. In both cases, the random distributions capture a distribution over a set of deterministic, ordinary differential equations. This should be seen in stark contrast to stochastic differential equations, where the randomness in the state space, i.e., diffusion, is modeled with a stochastic process. In comparison to \eqref{equation: dynamics model}, the latter is a time-varying disturbance added to the vector field. In that sense, our model still captures the {\em epistemic} uncertainty about our system dynamics, while an SDE model captures the intrinsic process noise, i.e., {\em aleatoric} uncertainty. While this reparametrization does not allow us to directly calculate $p_{D}(\dot{\mathcal{X}} | \mathcal{D}, \mathcal{T})$, we obtain a Gaussian distribution for the marginals $p_D(\dot{\bm{x}}_{\text{supp}} | \bm{x}_{\text{supp}})$. To retrieve $p_{D}(\dot{\mathcal{X}} | \mathcal{D}, \mathcal{T})$, we use the smoother model's predictive state posterior to obtain
\begin{align}
    p_{D}(\dot{\mathcal{X}} | \mathcal{D}, \mathcal{T}) &= \int p_D(\dot{\bm{x}}_{\text{supp}}, \bm{x}_{\text{supp}} | \mathcal{D}, \mathcal{T}) d \bm{x}_{\text{supp}}\\
    &\approx \int p_D(\dot{\bm{x}}_{\text{supp}} | \bm{x}_{\text{supp}}) p_S(\bm{x}_{\text{supp}} | \mathcal{T}, \mathcal{D})d \bm{x}_{\text{supp}}.
\end{align}
\parSpace

\subsection{Comparing Gradient Distributions via the Wasserstein Distance}
\parSpace
To compare and eventually match $p_{D}(\dot{\mathcal{X}} | \mathcal{D}, \mathcal{T})$ and $p_{S}(\dot{\mathcal{X}} | \mathcal{D}, \mathcal{T})$, we propose to use the Wasserstein distance \citep{kantorovich1939mathematical}, since it allows for an analytic, closed-form 
representation, and since it outperforms similar measures (like forward, backward and symmetric KL divergence) in our exploratory experiments. The squared type-2 Wasserstein distance gives rise to the term
\begin{equation}
    \label{equation: Wasserstein accurate}
    \mathbb{W}_2^2 \left[ p_{S}(\dot{\mathcal{X}} | \mathcal{D}, \mathcal{T}), p_{D}(\dot{\mathcal{X}} | \mathcal{D}, \mathcal{T}) \right]
    = \mathbb{W}_2^2 \left[ p_{S}(\dot{\mathcal{X}} | \mathcal{D}, \mathcal{T}), \mathbb{E}_{\bm{x}_{\text{supp}} \sim p_{\text{GP}}(\bm{x}_{\text{supp}} | \mathcal{D}, \mathcal{T})}\left[ p_D(\dot{\bm{x}}_{\text{supp}} | \bm{x}_{\text{supp}}) \right] \right]
\end{equation}
that we will later use to regularize the smoothing process.
To render the calculation of this regularization term computationally feasible, we introduce two approximations. First, observe that an exact calculation of the expectation in \Cref{equation: Wasserstein accurate}  requires mapping a multivariate Gaussian through the deterministic neural networks parametrizing $\bm{f}$ and $\bm{\Sigma}_D$ in \Cref{equation: dynamics model}. To avoid complex sampling schemes, we carry out a certainty-equivalence approximation of the expectation, that is, we evaluate the dynamics model on the posterior smoother mean $\bm{\mu}_{\text{S, supp}}$. As a result of this approximation, observe that both $p_{D}(\dot{\mathcal{X}} | \mathcal{D}, \mathcal{T})$ and $p_{S}(\dot{\mathcal{X}} | \mathcal{D}, \mathcal{T})$ become Gaussians. However, the covariance structure of these matrices is very different. Since we use independent GPs for different state dimensions, the smoother only models the covariance between the state values within the same dimension, across different time points. Furthermore, since $\bm{\epsilon}$, the random variable that captures the randomness of the dynamics across all time-points, is only $K$-dimensional, the covariance of $p_D$ will be degenerate. Thus, we do not match the distributions directly, but instead match the marginals of each state coordinate at each time point independently at the different supporting time points. Hence, using first marginalization and then the certainty equivalence, \Cref{equation: Wasserstein accurate} reduces to
\begin{align}
\label{equation: Wasserstein approximate}
\mathbb{W}_2^2 \left[ p_{S}(\dot{\mathcal{X}} | \mathcal{D}, \mathcal{T}), p_{D}(\dot{\mathcal{X}} | \mathcal{D}, \mathcal{T}) \right] 
&\approx \sum_{k=1}^K \sum_{i=1}^{|\dot{\mathcal{X}}|} \mathbb{W}_2^2 \left[ p_{S}(\dot{x}_{\text{supp}}^{(k)}(t_{\text{supp}, i}) | \mathcal{D}, \mathcal{T}), p_{D}(\dot{x}_{\text{supp}}^{(k)}(t_{\text{supp}, i}) | \mathcal{D}, \mathcal{T}) \right] \nonumber\\
&\mkern-90mu\approx \sum_{k=1}^K \sum_{i=1}^{|\dot{\mathcal{X}}|} \mathbb{W}_2^2 \left[ p_{S}(\dot{x}_{\text{supp}}^{(k)}(t_{\text{supp}, i}) | \mathcal{D}, \mathcal{T}), p_{D}(\dot{x}_{\text{supp}}^{(k)}(t_{\text{supp}, i}) | \bm{\mu}_{\text{S, supp}}) \right].
\end{align}
Conveniently, the Wasserstein distance can now be calculated analytically, since for two one-dimensional Gaussians $a \sim \mathcal{N}(\mu_a, \sigma_a^2)$ and $b \sim \mathcal{N}(\mu_b, \sigma_b^2)$, we have $\mathbb{W}_2^2[a, b] = (\mu_a-\mu_b)^2 + (\sigma_a - \sigma_b)^2$.
\parSpace

\subsection{Final Loss Function}
\parSpace
\looseness-1
As explained in the previous paragraphs, distributional gradient matching trains a smoother regularized by a dynamics model. Both the parameters of the smoother $\bm{\varphi}$, consisting of the trainable parameters of the GP prior mean $\bm{\mu}$, the feature map $\phi$, and the kernel $\mathcal{K}$, and the parameters of the dynamics model $\bm{\psi}$ are trained concurrently, using the same loss function. This loss consists of two terms, of which the regularization term was already described in \Cref{equation: Wasserstein approximate}. While this term ensures that the smoother follows the dynamics, we need a second term ensuring that the smoother also follows the data. To this end, we follow standard GP regression literature, where it is common to learn the GP hyperparameters by maximizing the marginal log likelihood of the observations, i.e. $\mathcal{L}_{\text{data}}$ \citep{Rasmussen2004}. Combining these terms, we obtain the final objective
\begin{equation*}
    \mathcal{L}(\bm{\varphi}, \bm{\psi}) \coloneqq 
    \mathcal{L}_{\text{data}}
    - \lambda  \cdot \sum_{k=1}^K \sum_{i=1}^{|\dot{\mathcal{X}}|} \mathbb{W}_2^2 \left[ p_{S}(\dot{x}_{\text{supp}}^{(k)}(t_{\text{supp}, i}) | \mathcal{D}, \mathcal{T}), p_{D}(\dot{x}_{\text{supp}}^{(k)}(t_{\text{supp}, i}) | \bm{\mu}_{\text{S, supp}}) \right].
\end{equation*}
\looseness -1 This loss function is a multi-criteria objective, where fitting the data (via the smoother) and identifying the dynamics model (by matching the marginals) regularize each other. 
In our preliminary experiments, we found the objective to be quite robust w.r.t. different choices of $\lambda$. In the interest of simplicity, we thus set it in all our experiments in \Cref{section: Experiments} to a default value of $\lambda = \frac{|\mathcal{D}|}{|\dot{\mathcal{X}}|}$, accounting only for the possibility of having  different numbers of supporting points and observations. One special case worth mentioning is $\lambda \rightarrow 0$, which corresponds to conventional sequential smoothing, where the second part would be used for identification in a second step, as proposed by \citet{pillonetto2010new}. However, as can be seen in \Cref{figure: ablation study}, the smoother fails to properly identify the system without any knowledge about the dynamics and thus fails to provide meaningful state or derivative estimates. Thus, especially in the case of sparse observations, joint training is strictly superior.
\parSpace

In its final form, unlike its pure Bayesian counterparts, $\operatorname{DGM}$ does not require any prior knowledge about the system dynamics. Nevertheless, if some prior knowledge is available, one could add an additional, additive term $\log(p(\bm{\psi}))$ to $\mathcal{L}(\bm{\varphi}, \bm{\psi})$. It should be noted however that this was not done in any of our experiments, and excellent performance can be achieved without.
\parSpace
\headSpace
\section{Experiments}
\headSpace
\label{section: Experiments}
We now compare $\operatorname{DGM}$ against state-of-the-art methods. In a first experiment, we demonstrate the effects of an overparametrized, simple dynamics model on the performance of $\operatorname{DGM}$ as well as traditional, MC-based algorithms $\operatorname{SGLD}$ (Stochastic Gradient Lengevin Dynamics, \citep{welling2011bayesian}) and $\operatorname{SGHMC}$ (Stochastic Gradient Hamiltonian Monte Carlo, \citep{chen2014stochastic}). We select our baselines based on the results of \citet{dandekar2020bayesian}, who demonstrate that both a variational approach and $\operatorname{NUTS}$ (No U-Turn Sampler, \citet{hoffman2014no}) are inferior to these two. Subsequently, we will investigate and benchmark the ability of $\operatorname{DGM}$ to correctly identify neural dynamics models and to generalize across different initial conditions. Since $\operatorname{SGLD}$ and $\operatorname{SGHMC}$ reach their computational limits in the generalization experiments, we compare against Neural ODE Processes ($\operatorname{NDP}$). Lastly, we will conclude by demonstrating the necessity of all of its components. For all comparisons, we use the julia implementations of $\operatorname{SGLD}$ and $\operatorname{SGHMC}$ provided by \citet{dandekar2020bayesian}, the pytorch implementation of $\operatorname{NDP}$ provided by \citet{norcliffe2021neural}, and our own JAX \citep{jax2018github} implementation of $\operatorname{DGM}$.
\parSpace

\subsection{Setup}
\parSpace
\looseness -1 We use known parametric systems from the literature to generate simulated, noisy trajectories. For these benchmarks, we use the two-dimensional {\em Lotka Volterra (LV)} system, the three-dimensional, chaotic {\em Lorenz (LO)} system, a four-dimensional {\em double pendulum (DP)} and a twelve-dimensional {\em quadrocopter (QU)} model. For all systems, the exact equations and ground truth parameters are provided in the \Cref{section: dataset description}. For each system, we create two different data sets. In the first, we include just one densely observed trajectory, taking the computational limitations of the benchmarks into consideration. In the second, we include many, but sparsely observed trajectories (5 for LV and DP, 10 for LO, 15 for QU). This setting aims to study generalization over different initial conditions.
\parSpace

\subsection{Metric} \looseness -1 We use the log likelihood as a metric to compare the accuracy of our probabilistic models. In the 1-trajectory setting, we take a grid of 100 time points equidistantly on the training trajectory. We then calculate the ground truth and evaluate its likelihood using the predictive distributions of our models. When testing for generalization, we repeat the same procedure for unseen initial conditions.
\parSpace

\subsection{Effects of Overparametrization}
\parSpace
\begin{wrapfigure}{ht}{0.33\textwidth}
\vspace{-0.5cm}
    \includegraphics[width=0.33\textwidth]{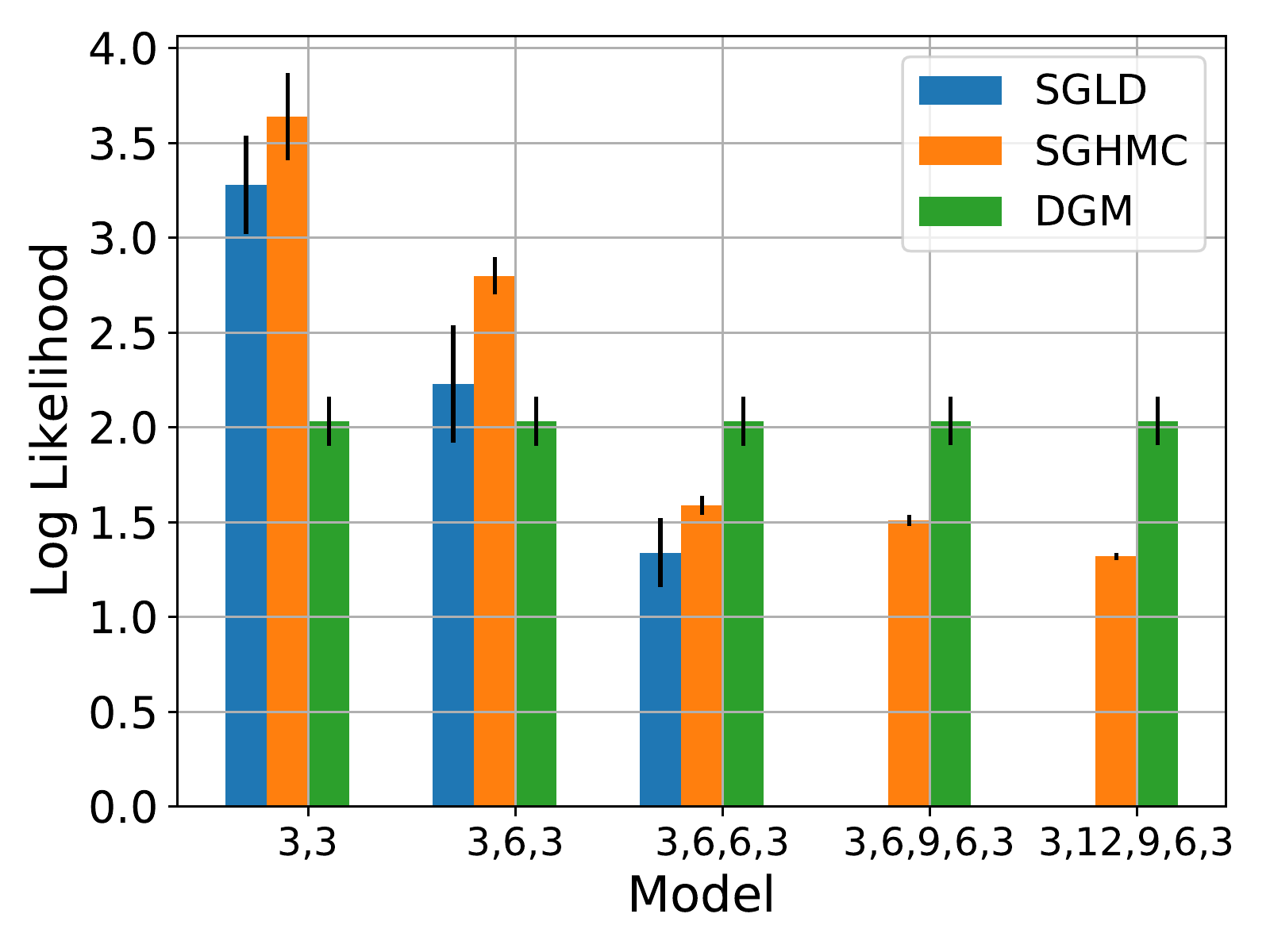}
    \vspace{-0.5cm}
    \caption{\looseness -1  $\operatorname{SGLD}$ does not converge for strongly overparametrized models, the performance of $\operatorname{SGHMC}$ deteriorates.  $\operatorname{DGM}$ is not noticeably affected.}
    \vspace{-0.2cm}
    \label{figure: overparametrization experiment}
\end{wrapfigure}
We first study a three-dimensional, linear system of the form $\dot{\bm{x}}(t) = \bm{A}\bm{x}(t)$, where $\bm{A}$ is a randomly chosen matrix with one stable and two marginally stable eigenvalues. For the dynamics model, we choose a linear Ansatz $\bm{f}(\bm{x}, \bm{\theta}) = \bm{B}\bm{x}(t)$, where $\bm{B}$ is parametrized as the product of multiple matrices. The dimension of the matrices of each factorization are captured in a string of integers of the form $(3,a_1, \dots, a_J, 3)$. For example, $(3,3)$ corresponds to $\bm{B}$ being just one matrix, while ($3, 6, 6, 3)$ corresponds to $\bm{B}=  \bm{B}_1 \bm{B}_2 \bm{B}_3$, with $\bm{B}_1 \in \mathbb{R}^{3 \times 6}$, $\bm{B}_2 \in \mathbb{R}^{6 \times 6}$ and $\bm{B}_3 \in \mathbb{R}^{6 \times 3}$. All of these models can be interpreted as linear neural networks, forming a simple case of the nonparametric systems we study later. Unlike general neural networks, the expressiveness of the Ansatz is independent of the number of parameters, allowing us to isolate the effects of overparametrization. In \Cref{figure: overparametrization experiment}, we show the mean and standard deviation of the log likelihood of the ground truth over $10$ different noise realizations. The exact procedure for one noise realization is described in the appendix, \Cref{section: Bayesian NODE training}). While $\operatorname{SGLD}$ runs into numerical issues after a medium model complexity, the performance of $\operatorname{SGHMC}$ continuously disintegrates, while $\operatorname{DGM}$ is unaffected. This foreshadows the results of the next two experiments, where we observe that the MC-based approaches are not suitable for the more complicated settings.
\parSpace

\begin{table}[!ht]
\centering
\captionof{table}{Log likelihood and prediction times of 100 ground truth sample points, with mean and standard deviation taken over 10 independent noise realizations, for neural ODEs trained on a single, densely sampled trajectory. }
\label{table: 1 trajectory training}
\scalebox{0.8}{
\begin{tabular}{@{}lcccccc@{}}
\toprule
& \multicolumn{3}{c}{Log Likelihood}  & \multicolumn{3}{c}{Prediction time [ms]} \\
\cmidrule(lr{1em}){2-4} \cmidrule{5-7}
& $\operatorname{DGM}$ & $\operatorname{SGHMC}$ & $\operatorname{SGLD}$ & $\operatorname{DGM}$   & $\operatorname{SGHMC}$ & $\operatorname{SGLD}$  \\
LV 1  & $\mathbf{1.96 \pm 0.21}$ & $1.36 \pm 0.0693$ & $1.03 \pm 0.0581$ & $\mathbf{0.68  \pm 0.04}$ & $14.98 \pm 0.23$  & $14.59 \pm 0.15$ \\
LO 1          & $\mathbf{-0.57 \pm 0.11}$& $-3.02 \pm 0.158$ & $-2.67 \pm 0.367$  & $\mathbf{0.99\pm 0.05}$   & $98.93. \pm 5.79$ & $105.03 \pm 12.22$\\
DP 1 & $\mathbf{2.13 \pm 0.14}$ & $1.88 \pm 0.0506$ & $1.85 \pm 0.0501$  & $\mathbf{1.31 \pm 0.05}$  & $10.60 \pm 0.21$  & $11.34 \pm 0.76$\\
QU 1    & $\mathbf{0.64 \pm 0.07}$ & $-5.00 \pm 1.36$  & NaN  & $\mathbf{3.76 \pm 0.12}$  & $24.68\pm 6.58$   & NaN \\
\bottomrule
\end{tabular}}
\end{table}

\subsection{Single Trajectory Benchmarks}
\parSpace
In \Cref{table: 1 trajectory training}, we evaluate the log-likelihood of the ground truth for the four benchmark systems, obtained when learning these systems using a neural ODE as a dynamics model (for more details, see \cref{section: implementation details of DGM}). Clearly, $\operatorname{DGM}$ performs the best on all systems, even though we supplied both $\operatorname{SGLD}$ and $\operatorname{SGHMC}$ with very strong priors and fine-tuned them with an extensive hyperparameter sweep (see \Cref{section: Bayesian NODE training} for more details). Despite this effort, we failed to get $\operatorname{SGLD}$ to work on Quadrocopter 1, where it always returned NaNs. This is in stark contrast to $\operatorname{DGM}$, which performs reliably without any pre-training or priors.
\parSpace

\subsection{Prediction speed}
\parSpace
To evaluate prediction speed, we consider the task of predicting $100$ points on a previously unseen trajectory. To obtain a fair comparison, all algorithms' prediction routines were implemented in JAX \citep{jax2018github}. Furthermore, while we used $1000$ MC samples when evaluating the predictive posterior for the log likelihood to guarantee maximal accuracy, we only used $200$ samples in \Cref{table: 1 trajectory training}. Here, $200$ was chosen as a minimal sample size guaranteeing reasonable accuracy, following a preliminary experiment visualized in \Cref{section: Bayesian NODE training}. Nevertheless, the predictions of $\operatorname{DGM}$ are 1-2 orders of magnitudes faster, as can be seen in \Cref{table: 1 trajectory training}. This further illustrates the advantage of relying on a smoother instead of costly, numerical integration to obtain predictive posteriors in the state space.
\parSpace
\vspace{-0.5cm}
\begin{table}[!ht]
    \begin{minipage}{0.55\textwidth}
        \centering
        \caption{Log likelihood of 100 ground truth sample points, with mean and
        covariance taken over 10 independent noise realizations, for neural ODEs trained on a multiple, sparsely sampled trajectory.The number following the system name denotes the number of trajectories in the training set.}
        \label{table: multitrajectory benchmarks}
        \scalebox{0.8}{
        \begin{tabular}{@{}lcc@{}}
        \toprule
        & \multicolumn{2}{c}{Log Likelihood}        \\
        \cmidrule{2-3}
                            & $\operatorname{DGM}$ & $\operatorname{NDP}$   \\
        LV 100   & $\mathbf{1.81 \pm 0.08}$      & $0.62 \pm 0.27$  \\
        LO 125          & $\mathbf{-2.18 \pm 0.76}$     & $-2.85 \pm 0.05$ \\
        DP 100 & $\mathbf{1.86 \pm 0.05}$      & $0.88 \pm 0.05$  \\
        QU 64    & $\mathbf{-0.54 \pm 0.36}$     & $-0.91 \pm 0.07$        \\
         \bottomrule
        \end{tabular}}
    \end{minipage}
    \hfill
    \begin{minipage}{0.45\textwidth}
    \centering
            \includegraphics[width=00.85\textwidth]{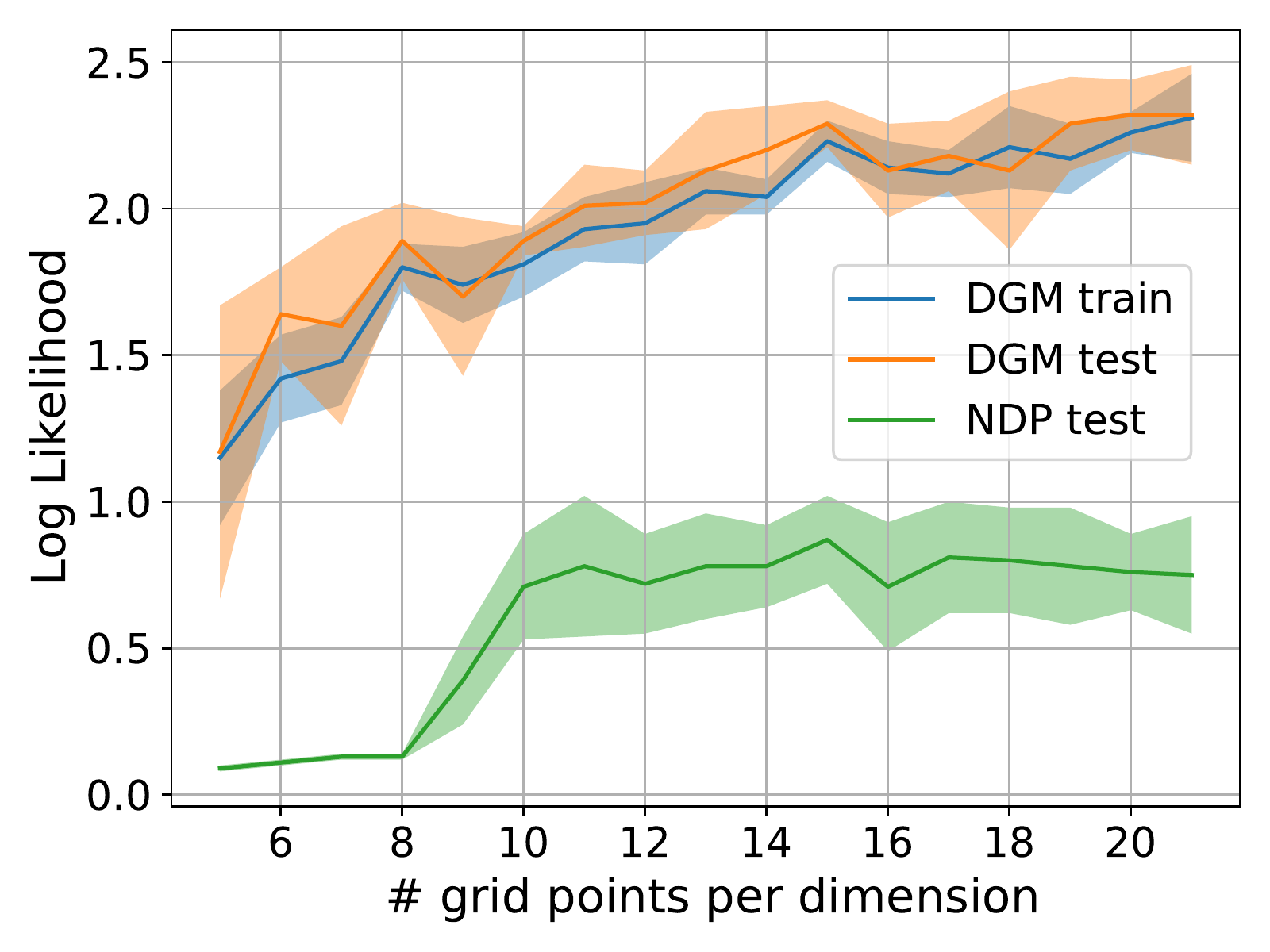}
            \captionof{figure}{Log likelihood of the ground truth for Lotka Volterra for increasing number of trajectories with 5 observations each.}
            \label{fig:learning curve}
    \end{minipage}
\end{table}

\subsection{Multi-Trajectory Benchmarks}
\parSpace
Next, we take a set of trajectories starting on an equidistant grid of the initial conditions. Each trajectory is then observed at $5$ equidistant observation times for LV and DP, and $10$ equidistant observation times for the chaotic Lorenz and more complicated Quadrocopter. We test generalization by randomly sampling a new initial condition and evaluating the negative log likelihood of the ground truth at $100$ equidistant time points. In \Cref{table: multitrajectory benchmarks}, we compare the generalization performance of $\operatorname{DGM}$ against $\operatorname{NDP}$, since despite serious tuning efforts, the MC methods failed to produce meaningful results in this setting. $\operatorname{DGM}$ clearly outperforms $\operatorname{NDP}$, a fact which is further exemplified in \Cref{fig:learning curve}. There, we show the test log likeliood for Lotka Volterra trained on an increasing set of trajectories. Even though the time grid is fixed and we only decrease the distance between initial condition samples, the dynamics model helps the smoother to generalize across time as well. In stark contrast, 
$\operatorname{NDP}$ fails to improve with increasing data after an initial jump.
\parSpace

\subsection{Ablation study}
\parSpace
We next study the importance of different elements of our approach via an ablation study on the Lorenz 125 dataset, shown in \Cref{figure: ablation study}. Comparing the two rows, we see that joint smoothing across trajectories is essential to transfer knowledge between different training trajectories. Similarly, comparing the two columns, we see that the dynamics model enables the smoother to reduce its uncertainty in between observation points.
\parSpace


\subsection{Computational Requirements}
\parSpace
For the one trajectory setting, all $\operatorname{DGM}$ related experiments were run on a Nvidia RTX 2080 Ti, where the longest ones took 15 minutes. The comparison methods were given 24h, on Intel Xeon Gold 6140 CPUs. For the multi-trajectory setting, we used Nvidia Titan RTX, where all experiments finished in less than 3 hours. A more detailed run time compilation can be found in \Cref{section: implementation details of DGM}.
Using careful implementation, the run time of $\operatorname{DGM}$ scales linearly in the number of dimensions $K$. However, since we use an accurate RBF kernel for all our experiments reported in this section, we have cubic run time complexity in $\sum_{m=1}^M N_m$. In principle, this can be alleviated by deploying standard feature approximation methods \citep{rahimi2007random, liu2020gaussian}. While this is a well known fact, we nevertheless refer the interested reader to a more detailed discussion of the subject in \Cref{section: scaling}.

\headSpace
\section{Related work}
\headSpace
\label{section: Related work}
\subsection{Bayesian Parameter Inference with Gaussian Processes}
\parSpace
The idea of matching gradients of a (spline-based) smoother and a dynamics model goes back to  the work of \citet{varah1982spline}. For GPs, this idea is introduced by \citet{calderhead2009accelerating}, who first fit a GP to the data and then match the parameters of the dynamics. \citet{dondelinger2013ode} introduce concurrent training, while \citet{gorbach2017scalable} introduce an efficient variational inference procedure for systems with a locally-linear parametric form. All these works claim to match the distributions of the gradients of the smoother and dynamics models, by relying on a product of experts heuristics. However, \citet{wenk2019fast} demonstrate that this product of experts in fact leads to statistical independence between the observations and the dynamics parameters, and that these algorithms essentially match \emph{point estimates} of the gradients instead. Thus, $\operatorname{DGM}$ is the first algorithm to actually match gradients on the level of distributions for ODEs. In the context of stochastic differential equations (SDEs) with constant diffusion terms, \citet{abbati2019ares} deploy MMD and GANs to match their gradient distributions. However, it should be noted that their algorithm treats the parameters of the dynamics model {\em deterministically} and thus, they can not provide the epistemic uncertainty estimates that we seek here. Note that our work is {\em not} related to the growing literature investigating SDE approximations of Bayesian Neural ODEs in the context of classification \citep{xu2021infinitely}. Similarly to \citet{chen2018neural}, these works emphasize learning a terminal state of the ODE used for other downstream tasks.
\vspace{-2mm}
\parSpace
\subsection{Gaussian Processes with Operator Constraints}
\parSpace
Gradient matching approaches mainly use the smoother as a proxy to infer dynamics parameters. This is in stark contrast to our work, where we treat the smoother as the main model used for prediction. While the regularizing properties of the dynamics on the smoother are explored by \citet{wenk2020odin}, \citet{jidling2017linearly} introduce an algorithm to incorporate linear operator constraints directly on the kernel level. Unlike in our work, they can provide strong guarantees that the posterior always follows these constraints. However, it remains unclear how to generalize their approach to the case of complex, nonlinear operators, potentially parametrized by neural dynamics models.
\parSpace

\subsection{Other Related Approaches}
\parSpace
\looseness -1 In some sense, the smoother is mimicking a probabilistic numerical integration step, but without explicitly integrating. In spirit, this approach is  similar to the solution networks used in the context of PDEs, as presented by \citet{raissi2019physics}, which however typically disregard uncertainty. In the context of classical ODE parameter inference, \citet{kersting2020differentiable} deploy a GP to directly mimic a numerical integrator in a probabilistic, differentiable manner. Albeit promising in a classical, parametric ODE setting, it remains unclear how these methods can be scaled up, as there is still the numerical integration bottleneck. Unrelated to their work, \citet{ghosh2021variational} present a variational inference scheme in the same, classical ODE setting. However, they still keep distributions over all weights of the neural network \citep{norcliffe2021neural}. A similar approach is investigated by \citet{dandekar2020bayesian}, who found it to be inferior to the MC methods we use as a benchmark. Variational inference was previously employed by \citet{yildiz2019ode2vae} in the context of latent neural ODEs parametrized by a Bayesian neural network, but their work mainly focuses on dimensionality reduction. Nevertheless, their work inspired a model called Neural ODE Processes by \citet{norcliffe2021neural}. This work is similar to ours in the sense that it avoids keeping distributions over network weights and models an ensemble of deterministic ODEs via a global context variable. Consequently, we use it as a benchmark in \Cref{section: Experiments}, showing that it does not properly capture epistemic uncertainty in a low data setting, which might be problematic for downstream tasks like reinforcement learning.
\parSpace

\headSpace
\section{Conclusion}
\headSpace
\label{section: Conclusion}
In this work, we introduced a novel, GP-based collocation method, that matches gradients of a smoother and a dynamics model on the distribution level using a Wasserstein loss. Through careful parametrization of the dynamics model, we manage to train complicated, neural ODE models, where state of the art methods struggle. We then demonstrate that these models are able to accurately predict unseen trajectories, while capturing epistemic uncertainty relevant for downstream tasks. In future work, we are excited to see how our training regime can be leveraged in the context of active learning of Bayesian neural ordinary differential equation for continuous-time reinforcement learning.

\section*{Acknowledgments}
This research was supported by the Max Planck ETH Center for Learning Systems. This project has
received funding from the European Research Council (ERC) under the European Union’s Horizon
2020 research and innovation programme grant agreement No 815943 as well as from the Swiss
National Science Foundation under NCCR Automation, grant agreement 51NF40 180545.
\bibliographystyle{apalike}
\bibliography{refs.bib}
\newpage
\appendix

\renewcommand*\contentsname{Contents of Appendix}
\etocdepthtag.toc{mtappendix}
\etocsettagdepth{mtchapter}{none}
\etocsettagdepth{mtappendix}{subsection}
\tableofcontents

\newpage
\section{Dataset description}
\label{section: dataset description}

In this section, we describe the datasets we use in our experiments.

\subsection{Lotka Volterra}
\label{subsection: Lotka Volterra}
The two dimensional Lotka Volterra system is governed by the parametric differential equations
\begin{align*}
    \frac{dx}{dt} &= \alpha x - \beta xy      \\
    \frac{dy}{dt} &= \delta xy - \gamma y,
\end{align*}
where we selected $(\alpha, \beta, \gamma, \delta) = (1, 1, 1, 1)$. These equations were numerically integrated to obtain a ground truth, where the initial conditions and observation times depend on the dataset. All observations were then created by adding additive, i.i.d. noise, distributed according to a normal distribution $\normal{0}{0.1^2}$.

\looseness-1
LV 1 consists of one trajectory starting from initial condition $(1, 2)$. The trajectory is observed at $100$ equidistant time points from the interval $(0, 10)$.

LV 100 consists of 100 trajectories. Initial conditions for these trajectories are located on a grid, i.e., 
\begin{align*}
\left\{ \left(\frac{1}{2} + \frac{i}{9}, \frac{1}{2} + \frac{j}{9}\right)\Big| i \in \{0, \dots, 9\}, j \in \{0, \dots, 9\} \right\}.    
\end{align*}
Each trajectory is then observed at $5$ equidistant time points from the interval $(0, 10)$, which leads to a total of $500$ observations.
\begin{figure}[H]
    \centering
    \includegraphics[width=1\linewidth]{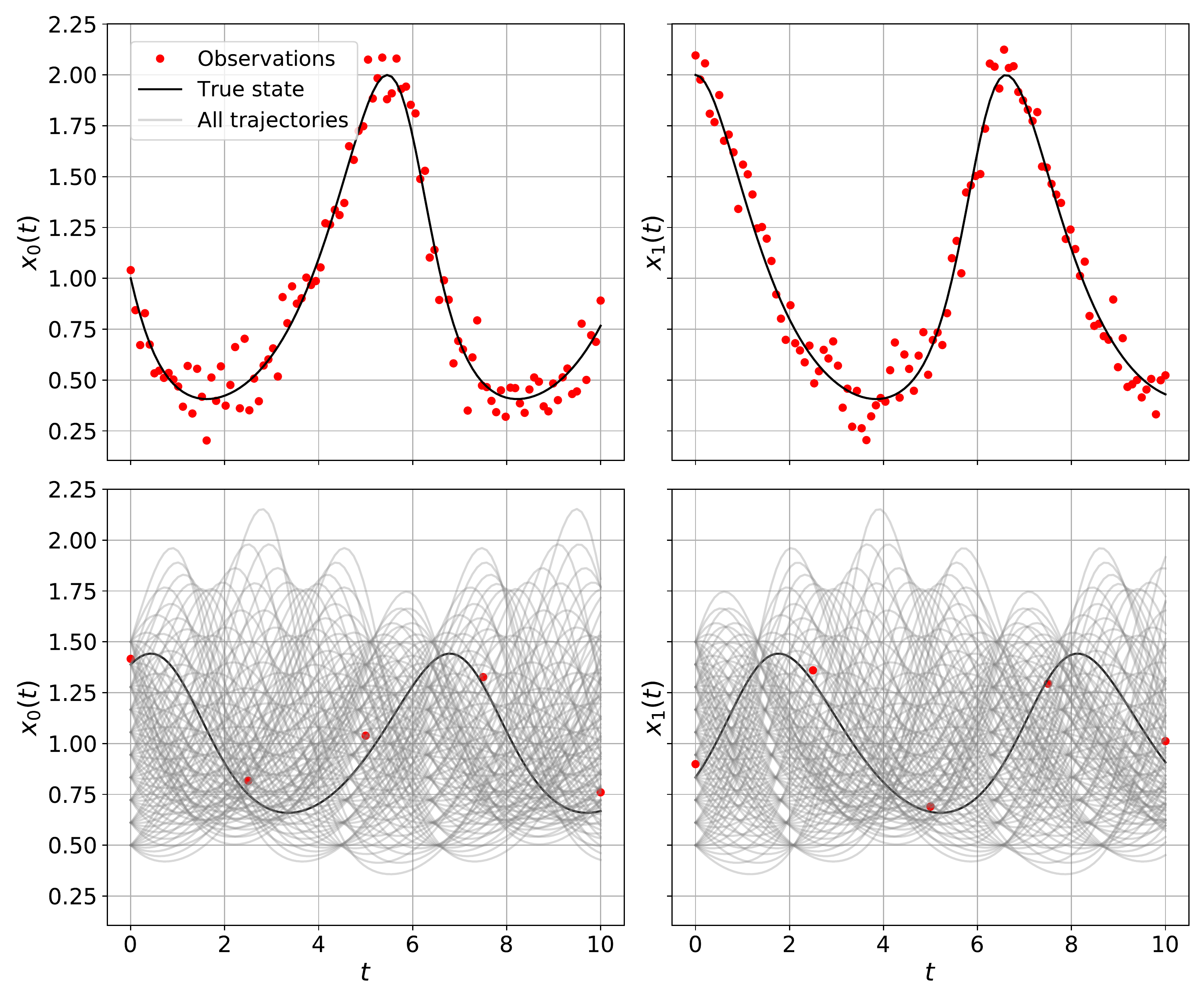}
    \caption{The first row represents the true states and all observations of LV 1 with random seed 0. In the second row we plot all ground truth trajectories from the dataset LV 100. One particular trajectory is highlighted in black, together with the corresponding observations of that trajectory (red dots).}
\end{figure}
To test generalization, we created $10$ new trajectories. The initial conditions of these trajectories were obtained by sampling uniformly at random on $[0.5, 1.5]^2$. To evaluate the log likelihood, we used $100$ equidistant time points from the interval $(0, 10)$.

\subsection{Lorenz}
\label{subsecttion: Lorenz}
The 3 dimensional, chaotic Lorenz system is governed by the parametric differential equations
\begin{align*}
&\frac{dx}{dt} = \sigma(y - x) \\
& \frac{dy}{dt} = x(\rho -z) - y \\
& \frac{dz}{dt} =  xy - \tau y,
\end{align*}
where we selected $(\sigma, \rho, \tau) = (10, 28, 8/3)$. These equations were numerically integrated to obtain a ground truth, where the initial conditions and observation times depend on the dataset. All observations were then created by adding additive, i.i.d. noise, distributed according to a normal distribution $\normal{0}{1}$.

LO 1 consists of one trajectory starting from initial condition $(-2.5, 2.5, 2.5)$. The trajectory is observed at $100$ equidistant time points from the interval $(0, 1)$.

LO 125 consists of 125 trajectories. Initial conditions for these trajectories are located on a grid, i.e., 
\begin{align*}
    \left\{ \left(-5 + \frac{5i}{2}, -5 + \frac{5j}{2}, -5 + \frac{5k}{2}\right)\Big| i \in \{0, \dots, 4\}, j \in \{0, \dots, 4\}, k \in \{0, \dots, 4\} \right\}.   
\end{align*}

Each trajectory is then observed on $10$ equidistant time points from the interval $(0, 1)$, which leads to a total of $1250$ observations.

\begin{figure}[ht]
    \centering
    \includegraphics[width=1\linewidth]{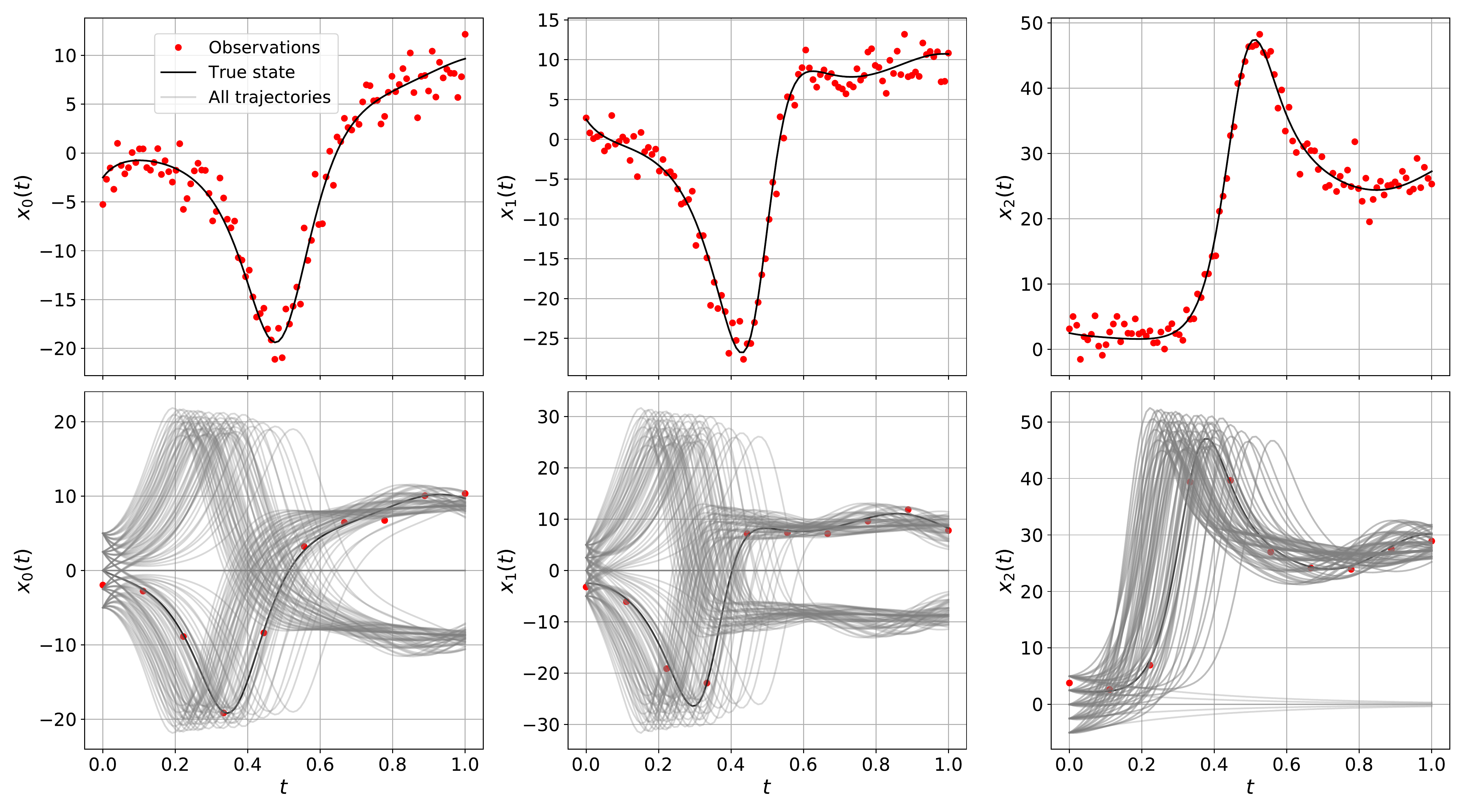}
    \caption{The first row represents the true states and all observations of LO 1 with random seed 0. In the second row we plot all ground truth trajectories from the dataset LO 125. One particular trajectory is highlighted in black, together with the corresponding observations of that trajectory (red dots).}
\end{figure}

To test generalization, we created $10$ new trajectories. The initial conditions of these trajectories were obtained by sampling uniformly at random on $[-5, 5]^3$. To evaluate the log likelihood, we used $100$ equidistant time points from the interval $(0, 1)$.

\newpage

\subsection{Double Pendulum}
\label{subsection: Double Pendulum}

The $4$ dimensional Double pendulum system is governed by the parametric differential equations

\begin{minipage}{.6\textwidth}
\begin{align*}
    {\dot \theta_1} &= \frac{6}{ml^2} \frac{ 2 p_{\theta_1} - 3 \cos(\theta_1-\theta_2) p_{\theta_2}}{16 - 9 \cos^2(\theta_1-\theta_2)} \\
{\dot \theta_2} &= \frac{6}{ml^2} \frac{ 8 p_{\theta_2} - 3 \cos(\theta_1-\theta_2) p_{\theta_1}}{16 - 9 \cos^2(\theta_1-\theta_2)}. \\
{\dot p_{\theta_1}} &=  -\tfrac{1}{2} m l^2 \left ( {\dot \theta_1} {\dot \theta_2} \sin (\theta_1-\theta_2) + 3 \frac{g}{l} \sin \theta_1 \right ) \\
{\dot p_{\theta_2}} &=  -\tfrac{1}{2} m l^2 \left ( -{\dot \theta_1} {\dot \theta_2} \sin (\theta_1-\theta_2) + \frac{g}{l} \sin \theta_2 \right ),
\end{align*}
\parbox{\textwidth}{where we selected $(g, m, l) = (9.81, 1, 1)$. In these equations, $\theta_1$ and $\theta_2$ represent the offset angles, while $p_{\theta_1}$ and $p_{\theta_1}$ represent the momentum of the upper and lower pendulum. These equations were numerically integrated to obtain a ground truth, where the initial conditions and observation times depend on the dataset. 
\parfillskip=0pt}
\vspace{-0.05cm}
\end{minipage}
\begin{minipage}{0.05\textwidth}
\,
\end{minipage}
\begin{minipage}{.34\textwidth}
    \centering
    \includegraphics[width=0.8\textwidth]{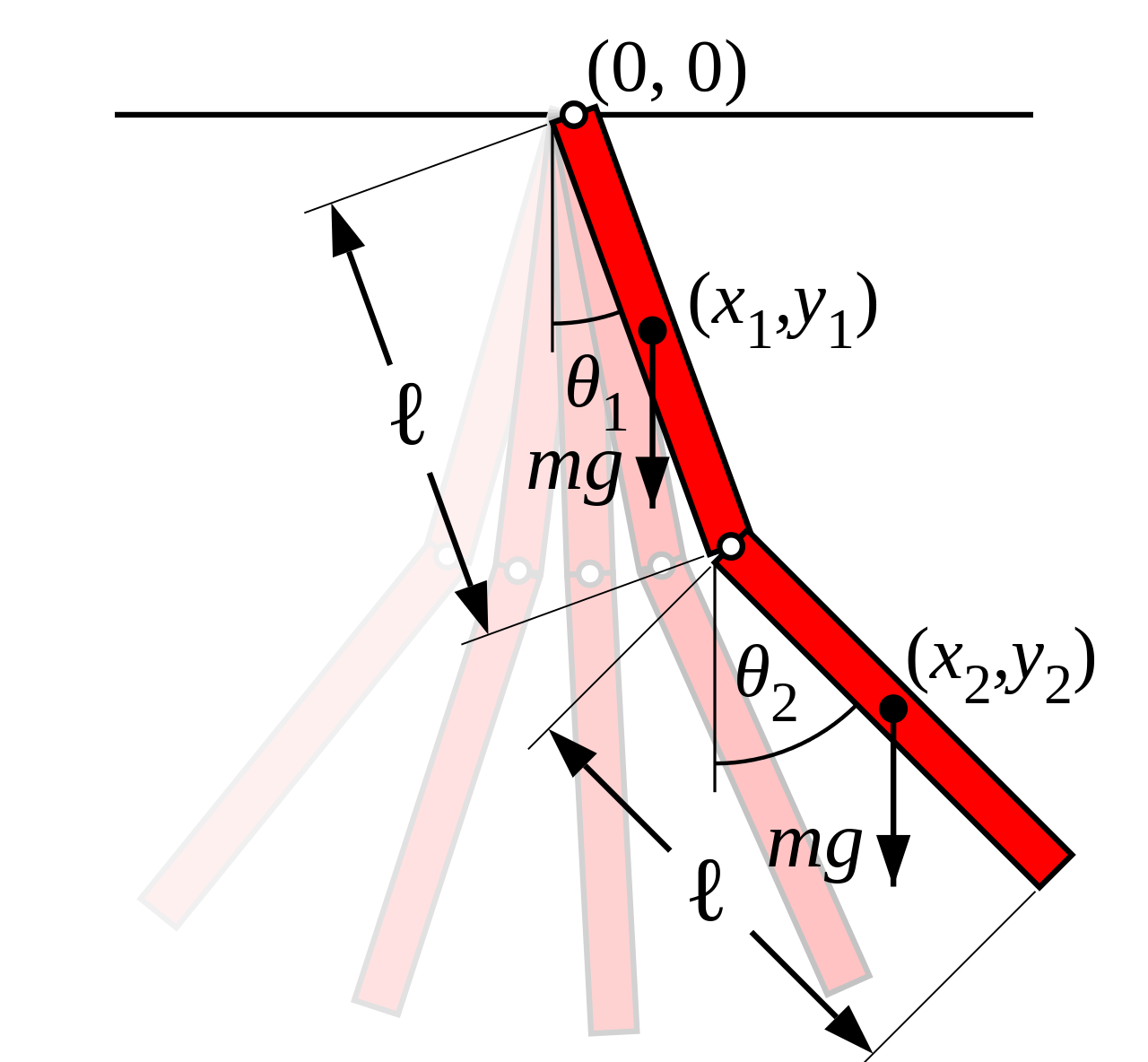}
    \captionof{figure}{Double Pendulum where both rods have equal length and mass.}
    \label{figure: double penudlum from wiki}
\end{minipage}
All observations were then created by adding additive, i.i.d. noise, distributed according to a normal distribution $\normal{0}{0.1^2}$.

DP 1 consist of one trajectory starting from initial condition $(-\pi/6, -\pi/6, 0, 0)$. The trajectory is observed at $100$ equidistant time points from the interval $(0,1)$.
 
 DP 100 consists of 100 trajectories. Initial conditions for these trajectories are located on a grid, i.e., 
 \begin{align*}
    \left\{\left(-\frac{\pi}{6} + \frac{\pi i}{27}, -\frac{\pi}{6} + \frac{\pi j}{27}, 0, 0\right)\Big| i \in \{0, \dots, 9\}, j \in \{0, \dots, 9\} \right\}. 
\end{align*}
Each trajectory is then observed at $5$ equidistant time points from the interval $(0, 1)$, which leads to a total of $500$ observations.

\begin{figure}[ht]
    \centering
    \includegraphics[width=1\linewidth]{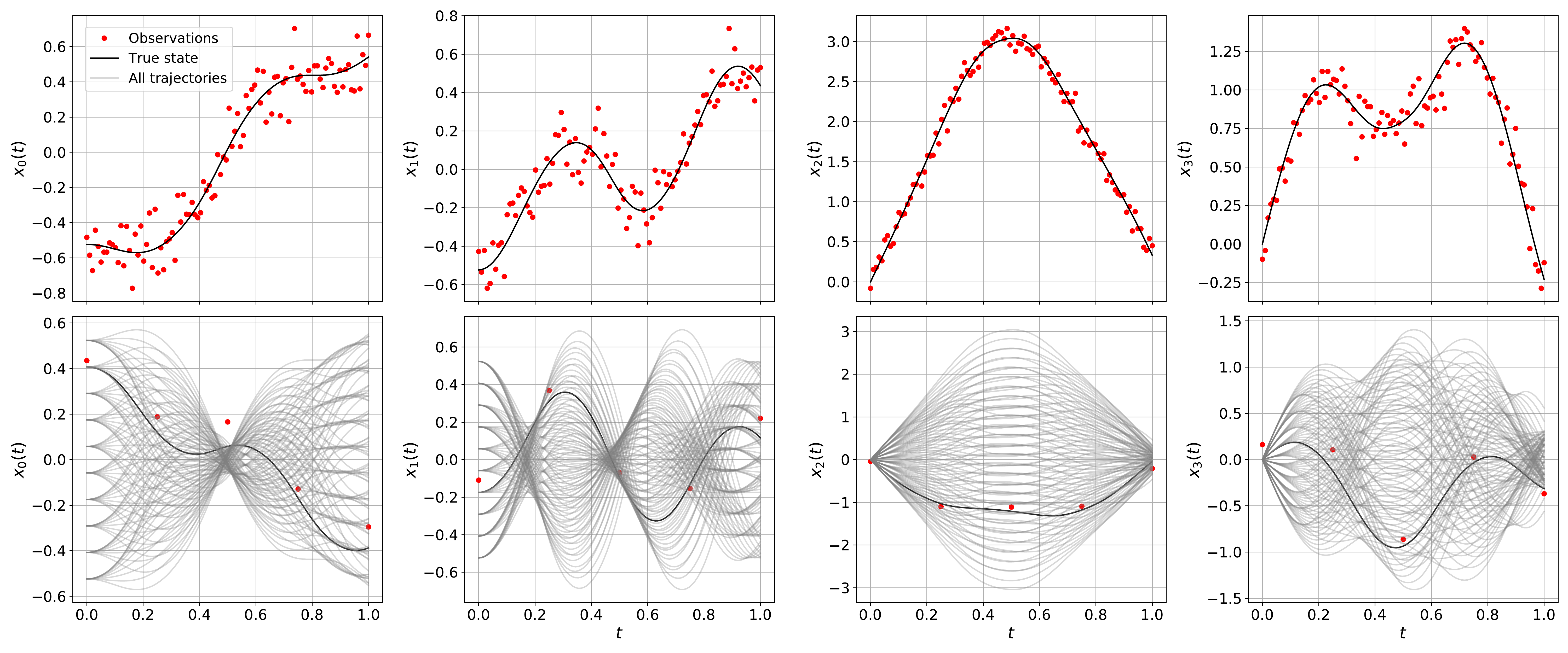}
    \caption{The first row represents the true states and all observations of DP 1 with random seed 0. In the second row we plot all ground truth trajectories from the dataset DP 100. One particular trajectory is highlighted in black, together with the corresponding observations of that trajectory (red dots).}
\end{figure}

To test generalization, we created $10$ new trajectories. The initial conditions of these trajectories were obtained by sampling uniformly at random on $[-\frac{\pi}{6}, \frac{\pi}{6}]^2 \times \{0\}^2$. To evaluate the log likelihood, we used $100$ equidistant time points from the interval $(0, 1)$.

\newpage
\subsection{Quadrocopter}
\label{subsection: Quadrocopter}
The $12$ dimensional Quadrocopter system is governed by the parametric differential equations
\begin{align*}
    \dot u &= -g \sin(\theta) + rv - qw \\
    \dot v &= g \sin(\phi) \cos(\theta) - ru + pw \\
    \dot w &= -F_z / m + g \cos(\phi)\cos(\theta) + qu - pv \\
    \dot p &= \left(L + (I_{yy} - I_{zz})qr\right)/ I_{xx} \\
    \dot q &=\left(M + (I_{zz} - I_{xx})pr\right) / I_{yy} \\
    \dot r &=(I_{xx}-I_{yy})pq / I_{zz} \\
    \dot \phi &= p + (q\sin(\phi) + r \cos(\phi))\tan(\theta)\\
    \dot \theta &= q \cos(\phi) - r \sin(\phi) \\
    \dot \psi &= (q \sin(\phi) + r \cos(\phi))\sec(\theta) \\
    \dot x &= \cos(\theta)\cos(\psi)u + (-\cos(\phi)\sin(\psi)+\sin(\phi)\sin(\theta)\cos(\psi))v + \\ &+(\sin(\phi)\sin(\psi) + \cos(\phi)\sin(\theta)\cos(\phi))w\\
    \dot y &=  \cos(\theta)\sin(\psi)u + (\cos(\phi)\cos(\psi)+\sin(\phi)\sin(\theta)\sin(\psi))v + \\ &+(-\sin(\phi)\cos(\psi) + \cos(\phi)\sin(\theta)\sin(\phi))w\\
    \dot z &= \sin(\theta)u - \sin(\phi)\cos(\theta)v - \cos(\phi)\cos(\theta)w,
\end{align*}
where
\begin{align*}
    F_z &= F_1 + F_2 + F_3 + F_4 \\
    L &= (F_2 + F_3)d_y - (F_1 + F_4)d_x \\
    M &= (F_1 + F_3)d_x - (F_2 + F_4)d_x.
\end{align*}
We fixed the input control forces to $(F_1, F_2, F_3, F_4) = (0.496 , 0.495 , 0.4955, 0.4955)$ (not to be inferred) and selected $(m, I_{xx}, I_{yy}, I_{zz}, d_x, d_y, g) = (0.1   , 0.62  , 1.13  , 0.9   , 0.114 , 0.0825, 9.85)$. These equations were numerically integrated to obtain a ground truth, where the initial conditions and observation times depend on the dataset. All observations were then created by adding additive, i.i.d. noise, distributed according to a normal distribution $\normal{0}{\Sigma}$, where
\begin{align*}
    \Sigma = \diag(1, 1, 1, 0.1, 0.1, 0.1, 1, 0.1, 0.1, 5, 5, 5).
\end{align*}

QU 1 consists of one trajectory starting from initial condition $(0,0,0,0,0,0,0,0,0,0,0,0)$. The trajectory is observed at $100$ equidistant time points from the interval $(0, 10)$.

QU 64 consists of $64$ trajectories. Initial conditions for these trajectories are located on a grid, i.e., 
\begin{align*}
        \left\{ \left(0,0,0,0,0,0,-\frac{\pi}{18} + \frac{\pi i}{27}, -\frac{\pi}{18} + \frac{\pi j}{27}, -\frac{\pi}{18} + \frac{\pi k}{27},0,0,0\right)\Big| (i, j, k) \in \{0, \dots, 4\}^3 \right\}.   
\end{align*}
Each trajectory is then observed at $15$ equidistant time points from the interval $(0, 10)$, which leads to a total of $960$ observations.

\begin{figure}[ht]
    \centering
    \includegraphics[width=0.8\linewidth]{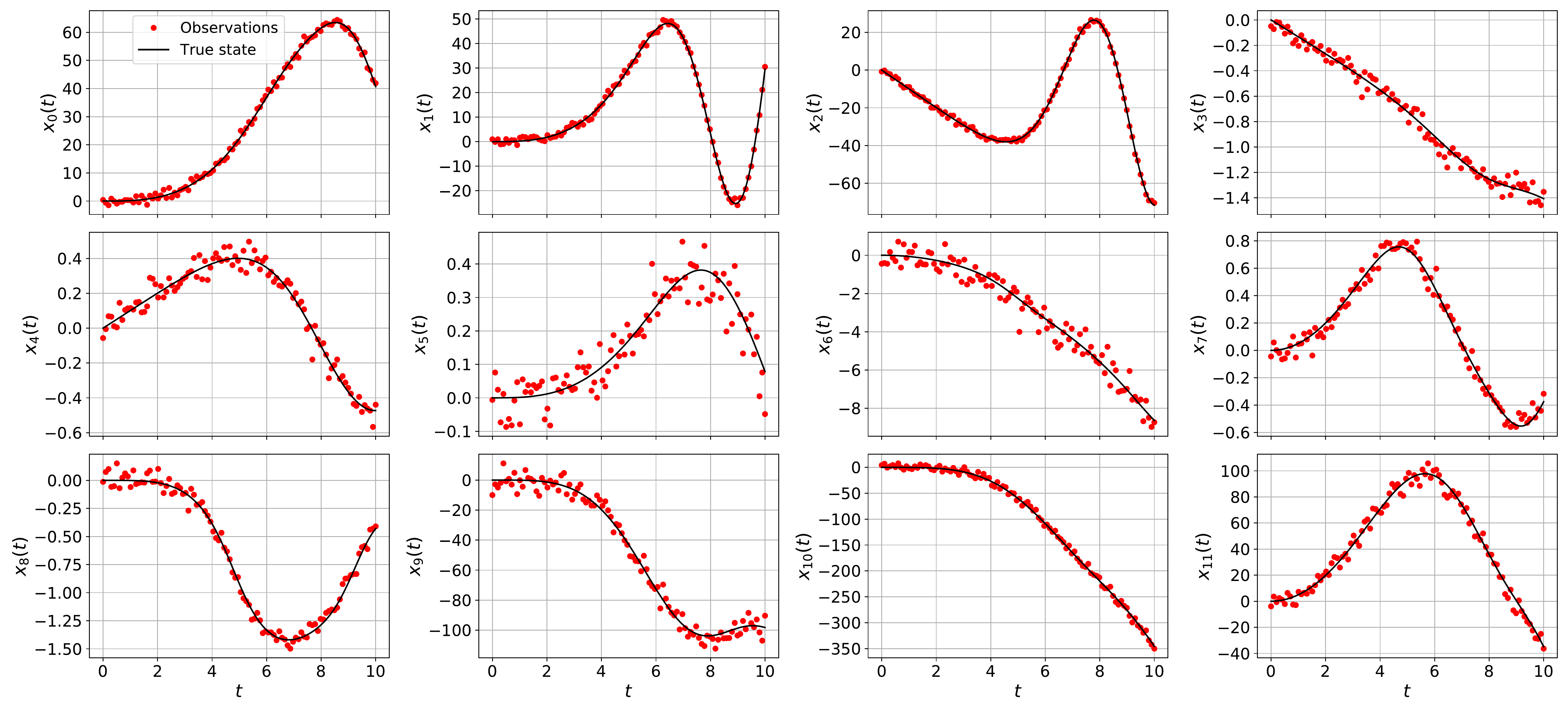}
    \caption{Visualization showing the true states and all observations of QU 1 with random seed 0.}
\end{figure}
\begin{figure}[ht]
    \centering
    \includegraphics[width=0.8\linewidth]{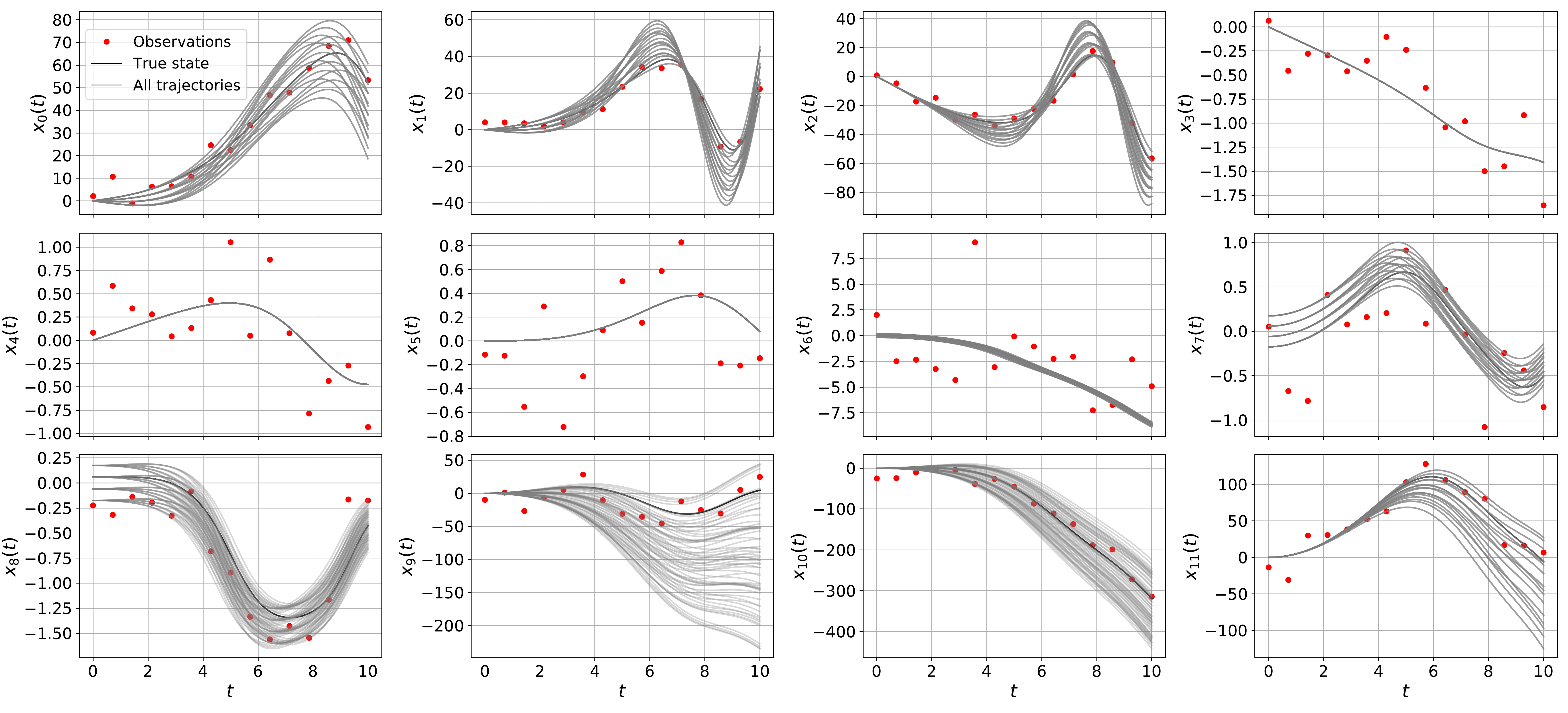}
    \caption{Visualization showing all ground truth trajectories from the dataset QU 64. One particular trajectory is highlighted in black, together with the corresponding observations of that trajectory (red dots).}
\end{figure}

To test generalization, we created $10$ new trajectories. The initial conditions of these trajectories were obtained by sampling uniformly at random on $\{0\}^6 \times [-\frac{\pi}{18}, \frac{\pi}{18}]^2 \times \{0\}^3$. To evaluate the log likelihood, we used $100$ equidistant time points from the interval $(0, 10)$.

\newpage
\section{Implementation details of $\operatorname{DGM}$}
\label{section: implementation details of DGM}

In this section we discuss all implementation details of $\operatorname{DGM}$. As described in \Cref{section: Distributional GM Key Ideas}, $\operatorname{DGM}$ consists of a smoother and a dynamics model. At training time, the dynamics model is then used to regularize the smoother via the squared type-2 Wasserstein distance.

\subsection{Dynamics model}
\label{subsection: dynamics model}
The role of the dynamics model is to map every state $\bm{x}$ to a distribution over derivatives, which we decide to parameterize as $\normal{\bm{f}(\bm{x}, \bm \psi)}{\bm \Sigma_D(\bm{x}, \bm{\psi})}$. In the paper we focused in the case where we do not have any prior knowledge about the dynamics and we model both $\bm f$ and $\bm \Sigma_D$ with neural network. Nevertheless, if some prior knowledge about the dynamics $\bm{f}$ is available, it can be used to inform the mean function $\bm f$ (potentially also covariance $\bm \Sigma_D$) appropriately. In particular, if the parametric form of the dynamics is known, we can use them as a direct substitute for $\bm{f}$. This is the special case of ODE-parameter inference, which we investigate empirically in \Cref{section: additional experiments}. Next we present implementation details of both non-parametric (i.e. $\bm{f}$ given by a neural network) and parametric (i.e. $\bm{f}$ given by some parametric form) dynamics models.

\paragraph{Non-parametric dynamics}In the non-parametric case, we model both the dynamics' mean function $\bm \mu_D$ and covariance function $\bm \Sigma_D$ with a neural networks. Across all experiments, we choose a simple 3-layer neural network with 20, 20 and $2n$ nodes, where $n$ denotes the number of state dimensions of a specific system. After each layer, we apply the sigmoid activation function, except for the last one. The first $n$ output nodes represent the mean function. Here, no activation function is necessary for the last layer. The second $n$ output nodes are used to construct the diagonal covariance $\bm \Sigma_D$. To ensure positivity, we use $x \mapsto \log(1+\exp(x))^2$ as an activation function on the last $n$ nodes of the last layer.

\paragraph{Parametric dynamics}In the parametric case, we model $\bm \mu_D$ using the parametric form of the vector field. Across all experiments, we choose a simple 3-layer neural network with 10, 10 and $n$ nodes, where $n$ denotes the number of state dimensions of a specific system. After each lyer, we apply the sigmoid activation function, except for the last one. The $n$ nodes are then used to construct the diagonal covariance $\bm \Sigma_D$. To ensure positivity, we use $x \mapsto \log(1+\exp(x))^2$ as an activation function on the last layer.

\subsection{Smoother model}
\label{subsection: smoother model}
The role of the smoother model is to map every tuple $(\bm{x}(0), t)$ consisting of initial condition $\bm{x}(0)$ and time $t$ to $\bm{x}(t)$, which is the state at time $t$ of a trajectory starting at $\bm{x}(0)$ at time $0$. In the paper, we model the smoother using a Gaussian process with a deep mean function $\bm{\mu}$ and a deep feature map $\phi$. Both of them take as input the tuple $(\bm{x}(0), t)$. This tuple is then mapped through a dense neural network we call core. For all experiments, we chose a core with two layers, with 10 and 5 hidden nodes and sigmoid activation on both. The output of the core is then fed into two linear heads. The head for $\bm{\mu}$ builds a linear combination of the core's output to obtain a vector of the same shape as $\bm{x}(t)$. The head for $\phi$ builds a linear combination of the core's output to obtain a vector of length $3$, the so called features. These features are then used as inputs to a standard RBF kernel with ARD \citep{Rasmussen2004}. For each state dimension, we keep a separate $\phi$-head, as well as separate kernel hyperparameters. However, the core is shared across dimensions, while $\bm{\mu}$ is directly introduced as multidimensional.

In the paper, we set the variance of the RBF to $1$ and learned the lengthscales together with all other hyperparameters. However, due to the expressiveness of the neural network, the lengthscales are redundant and could easily be incorporated into the linear combination performed by the head. Thus, in the scaling experiments, we fix the lengthscale to one and approximate the RBF kernel with a feature expansion, as detailed in \Cref{section: scaling}.

\subsection{Evaluation metric}
\label{subsection: evaluation metric}
To evaluate the quality of our models' predictions, we use the log likelihood. To obtain the log likelihood, we first use the model to predict the mean and standard deviation at 100 equidistant times. Then we calculate the log likelihood of the ground truth for every predicted point. We take the mean over dimensions, over times, and over trajectories. When reporting the training log likelihood, as done e.g. in \Cref{table: 1 trajectory training}, we use the training trajectories for evaluation. When reporting the generalization log likelihood, as done e.g. in \Cref{table: multitrajectory benchmarks}, we use $10$ unseen trajectories. This evaluation is then repeated for $10$ different , meaning that we retrain the model $10$ times on a data set with the same ground truth, but a different noise realization. We then report the mean and standard deviation of the log likelihood across these repetitions.

\paragraph{Weight decay} To prevent overfitting, we use weight decay on the parameters of both the dynamics and the smoother neural networks. We denote by $wd_D$ the weight decay parameter of the dynamics model, and $wd_S$ the weight decay parameters of the smoother model. While we keep the $wd_S$ constant during all three phases of training, we gradually increase $wd_D$ from $0$ to its final value, which is the same as $wd_S$. The increase follows a polynomial schedule with power $0.8$.

\subsection{Training details}
\label{subsection: training details}
The training of $\operatorname{DGM}$, i.e. optimizing \Cref{equation: Simplified Loss}, can be split into three distinct phases: \emph{transition}, \emph{training}, and \emph{fine-tuning}. In the \emph{transition} phase we gradually increase the value of both $\lambda$ and the weight decay regularization parameter of the dynamics $(wd_D)$ from 0 to its final value. When these parameters reach their final value, we reach the end of the transition phase and start the \emph{training} phase. In this phase, all optimization parameters are left constant. It stops when the last $1000$ steps are reached. Then, the \emph{fine-tune} phase starts, where we decrease learning rate to $0.01$. The gradual increase of $\lambda$ and $wd_D$ follows polynomial schedule with power $0.8$. As an optimizer, we use Adam.

\paragraph{Supporting points} The selection of the supporting points in $\mathcal{T}$ is different for data sets consisting of one or multiple trajectories. If there is only one trajectory in the dataset, we match the derivatives at the same places where we observe the state. If there are multiple trajectories, we match the derivatives at $30$ equidistant time points on each training trajectory.

\begin{figure}[ht]
    \centering
    \includegraphics[width=1.0\linewidth]{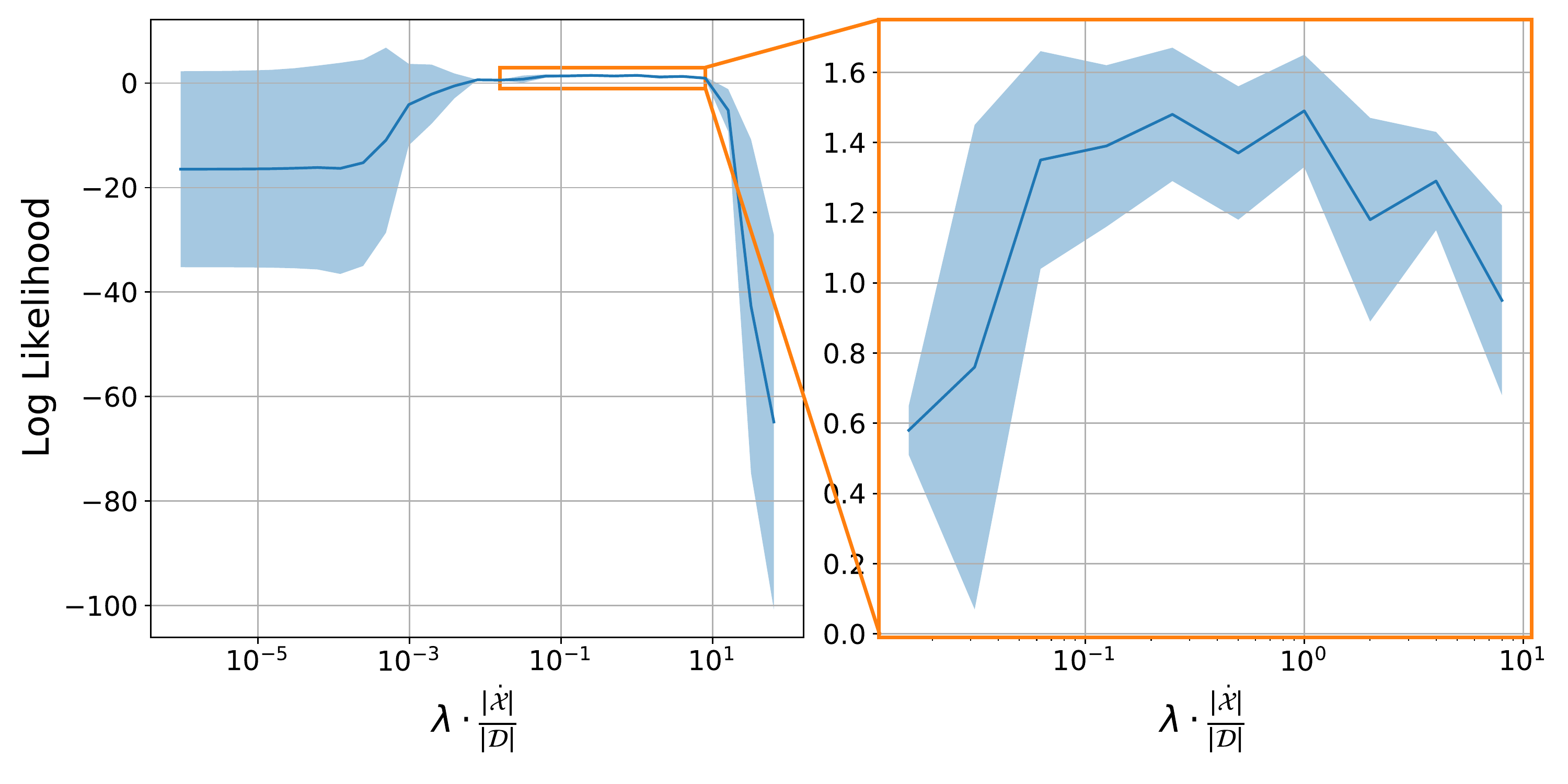}
    \caption{\looseness-1 If $\lambda$ is too small, the dynamics model does not regularize the smoother sufficiently, the model overfits to the data and the test log likelihood score is worse. If $\lambda$ is too large, the observation term gets dominated, the model underfits and the log likelihood score on the test data is worse. Empirically, we found that we achieve the best log likelihood on the test data with $\lambda = |\mathcal{D}|/|\dot{\mathcal{X}}|$.}
    \label{figure: lambda selection}
\end{figure}

\paragraph{Selection of \texorpdfstring{$\lambda$}{lambda}} The loss of \Cref{equation: Simplified Loss} is a multi-objective optimization problem with a trade-off parameter $\lambda$. Intuitively, if $\lambda$ is too small, the model only tries to fit the data and neglects the dynamics. On the other hand, with too large $\lambda$, the model neglects the data fit and only cares about the dynamics. In \Cref{figure: lambda selection} we show a plot of log likelihood score on the $10$ test trajectories of the LV 100 dataset with varying $\lambda$. We train the model for $\lambda \cdot |\dot{\mathcal{X}}|/|\mathcal{D}| \in \{2^i | i=-20, \ldots, 6\}$. To estimate the robustness of the experiment, we show the mean and standard deviation over $5$ different noise realizations.

\paragraph{Parameter selection}To search for the best performing parameters we performed sweep over learning rate value $lr$ in the transition and training phase and over the weight decay parameters $wd_S$. For $lr$, we considered the values $0.02, 0.05$ and $0.1$. For $wd_S$, we considered $0.1, 0.5$ and $1.0$.

\paragraph{Training time} We did not optimize our hyperparameters and code for training time. Nevertheless,  we report the length of each phase and the total time in the \Cref{table: training times}.
\begin{table}[ht]
\centering
\caption{Number of steps for each training phase and total training time for different datasets. For the times, we report mean $\pm$ standard deviation over 10 different random seeds.}
\label{table: training times}
\begin{tabular}{@{}lllll@{}}
\toprule
       & Transition & Training & Fine-Tuning & Time{[}s{]} \\ \midrule
LV 1   & 1000       & 0         & 1000      &      $329 \pm 15$       \\
LO 1   & 1000       & 0         & 1000      &       $399\pm 6$      \\
DP 1   & 1000       & 1000      & 1000      &    $535 \pm 42$         \\
QU 1   & 1000       & 2000      & 1000      &   $1121 \pm 41$          \\
LV 100 & 1000       & 0         & 1000      & $408 \pm 8$             \\
LO 125 & 1000       & 0         & 1000      & $753 \pm 8$            \\
DP 100 & 5000       & 4000      & 1000      &   $2988 \pm 261$          \\
QU 64  & 6000       & 3000      & 1000      &     $8387 \pm 36$        \\ \bottomrule
\end{tabular}
\end{table}
\newpage
\section{Bayesian NODE training}
\label{section: Bayesian NODE training}

In this section, we describe the specifics of the experiments and the implementation of $\operatorname{SGLD}$ and $\operatorname{SGHMC}$, the Bayesian integration benchmarks used in the paper. For all experiments, we use a slightly changed version of the code provided by \citet{dandekar2020bayesian}, which is written in Julia \citep{Julia-2017}.

\subsection{Effects of Overparametrization}
\label{subscetion: effects of overparameterization}
Here we provide further details regarding the experiment presented in \Cref{figure: overparametrization experiment}. For the ground truth dynamics $\dot{\bm{x}} = \bm{A} \bm{x}$, we selected the matrix $\bm{A}$ such that it has $1$ stable and $2$ marginally stable modes. The eigenvalue of the stable mode is selected uniformly at random from $[-0.5, -0.1]$. To create marginally stable modes, we create a block $\bm C \in \mathbb{R}^{2\times 2}$, where its components are sampled i.i.d. uniformly at random from $[0, 1]$. The marginally stable part is then created as $\bm A$ as $\frac{\pi}{2\rho(\bm C - \bm C^\top)}\left(\bm C - \bm C^\top\right) $, where $\rho(.)$ denotes the spectral radius. Using the spectral radius in the normalization ensures that the period of the mariginally stable mode is bounded with $\pi/2$. We selected the initial condition for the trajectory uniformly at random from the unit sphere in $\R^3$. We evolved the trajectory on the time interval $(0, 10)$ and observed 100 noisy observations, where every ground truth value was perturbed with additive, independent Gaussian noise, drawn from $\normal{0}{0.1^2}$.

While $\operatorname{DGM}$ performed without any pre-training, we observed that both $\operatorname{SGLD}$ and $\operatorname{SGHMC}$ struggle without narrow priors. To obtain reasonable priors, we followed the following procedure:

First, we pretrain every set of matrices $\bm B_1, \dots, \bm B_k$ on the ground truth, such that the
\begin{align*}
    \sum_{i=1}^{100}\norm{\dot{\bm{x}}(t_i) - \prod_{j=1}^{k} \bm{B}_j \bm{x}(t_i)}_2^2 \le 10 ^{-5},
\end{align*}
where $t_i$ are times at which we observed the state. We selected the prior as Gaussian centered around the pretrained parameters. The standard deviation was chosen such that the standard deviation of the product of the matrices $\prod_{j=1}^k\bm B_j$  stays at $0.01$, independent of the number of matrices used. For $\operatorname{SGHMC}$,  we selected learning rate $1.5 \times 10^{-7}$ and momentum decay $0.1$ as hyperparameters. For $\operatorname{SGLD}$,we chose the hyperparameters $a = 0.001, b = 1$ and $\gamma = 0.9$, which $\operatorname{SGLD}$ then uses to calculate a polynomial decay $a(b + k)^{-\gamma}$ (at step $k$) for its learning rate. For sampling we used $5$ chains with $20000$ samples on each chain. The last $2000$ samples of each chain were used for evaluation. With this setting we ensured that the $r$-hat score was smaller than 1.1.

\subsection{Finetuning for Benchmark Systems}
\label{subsection: hyperparameters selection}
Both $\operatorname{SGLD}$ and $\operatorname{SGHMC}$ require hyperparameters, that influence their performance quite strongly. In this subsection, we explain all the tuning steps and requirements we deployed for these algorithms to produce the results shown in the main paper. For both algorithms, we used a setup of $6$ chains, that were sampled in parallel, with $10000$ samples per chain, where the final $2000$ samples were taken as predictions. These numbers were selected using the r-hat value to determine if the chains had sufficiently converged.

\paragraph{Hyperparameters of $\operatorname{SGLD}$} For $\operatorname{SGLD}$, we additionally searched over the hyperparameters $a$, $b$, and $\gamma$. All three are used to calculate the learning rate of the algorithm. These parameters were chosen by evaluating the log likelihood of the ground truth on a grid, which was given by the values $a\in \{0.0001, 0.001, 0.005, 0.01, 0.05, 0.1\}$, $b\in\{0.3, 0.6, 1.0, 1.5 , 2\}$ and $\gamma \in \{0.5001, 0.55, 0.6, 0.7, 0.8, 0.99\}$. Clearly, using the log likelihood of the ground truth to tune the hyperparameters overestimates the performance of the algorithm, but it provides us with an optimistic estimate of its real performance in practice. For computational reasons, we only performed a grid search on the one trajectory experiments, and then reused the same hyperparameters on the multi-trajectory experiments. All hyperparameters are shown in \Cref{table: hyperparameters}.

\paragraph{Hyperparameters of $\operatorname{SGHMC}$} For $\operatorname{SGHMC}$, we additionally searched over the hyperparameters the learning rate and the momentum decay. Again, these parameters were chosen by evaluating the log likelihood of the ground truth on a grid, where learning rate was chosen from the set $\{1\mathrm{e}{-8}, 5\mathrm{e}{-8}, 1.5\mathrm{e}{-7}, 5\mathrm{e}{-7}, 1\mathrm{e}{-6}, 5\mathrm{e}{-6}, 1\mathrm{e}{-5}, 5\mathrm{e}{-5}, 1\mathrm{e}{-4}, 5\mathrm{e}{-4}\}$ and momentum decay was chosen from the set $\{0.0001, 0.001, 0.05, 0.1, 0.5, 1, 5\}$. Since we used the log likelihood of the ground truth again to tune the hyperparameters, we overestimate the performance of the algorithm and obtain thus an optimistic estimate of its real performance in practice. For computational reasons, be only performed a grid search on the one trajectory experiments, and then reused the same hyperparameters on the multi-trajectory experiments. All hyperparameters are shown in \Cref{table: hyperparameters}.

\begin{table}[!h]
\caption{Hyperparameters with best performance evaluated on the likelihood of the ground truth. The hyperparameters are different for the parametric (p) and the non-parametric (n) dynamics models, which is indicated with the last letter.}
\label{table: hyperparameters}
\centering
\begin{tabular}{lccccc}
\toprule
     & \multicolumn{3}{c}{$\operatorname{SGLD}$}&\multicolumn{2}{c}{$\operatorname{SGHMC}$} \\
     \cmidrule(r{1em}){2-4} \cmidrule{5-6}
     & $a$       & $b$    & $\gamma$            & learning rate   & momentum decay   \\
Lotka Volterra p & $1\mathrm{e}{-3}$    & $2   $ & $0.5001 $           & $5\mathrm{e}{-7}$          & $0.1  $          \\
Lorenz p & $0.001$   & $1.5 $ & $0.5001 $           & $1\mathrm{e}{-5}$          & $0.05 $          \\
Double Pendulum p & $0.1   $  & $0.3 $ & $0.5001 $           & $1\mathrm{e}{-6}$          & $0.1  $          \\
Quadrocopter p & $0.0001$  & $1.5 $ & $0.7    $           & $5\mathrm{e}{-7}$          & $0.5  $          \\
Lotka Volterra n & $0.005 $  & $1.5 $ & $0.7    $           & $5\mathrm{e}{-7}$          & $0.5  $          \\
Lorenz n & $0.001 $  & $1.5 $ & $0.55   $           & $5\mathrm{e}{-6}$          & $0.1  $          \\
Double Pendulum n & $0.01  $  & $2   $ & $0.55   $           & $1\mathrm{e}{-6}$          & $0.05 $          \\
Quadrocopter n & F         & F      & F                   & $5\mathrm{e}{-7}$          & $0.05 $ \\
\bottomrule
\end{tabular}
\end{table}

\paragraph{Choice of Priors for $\operatorname{SGLD}$ and $\operatorname{SGHMC}$} Since $\operatorname{SGLD}$ and $\operatorname{SGHMC}$ are both Bayesian methods, they need to be supplied with a prior. As we discovered in our experiments, this prior plays a crucial role in the stability of the algorithm. In the end, we did not manage to get them to converge without some use of ground truth. In particular, if the priors were not chosen narrowly around some ground truth, the algorithms just returned NaNs, since their integration scheme runs into numerical issues. For the parametric case shown in \Cref{section: additional experiments}, where we assume access to the true parametric form of the system, we thus chose narrow priors around the ground truth of the parameter values, that were used to create the data set. For Lotka Volterra, we chose a uniform distribution around the ground truth $\pm 0.5$, i.e. $\theta_i \sim \textrm{Uniform}[0.5, 1]$ for all components of $\bm{\theta}$. For Lorenz, we chose $\alpha \sim \textrm{Uniform}[8, 12]$, $\beta \sim \textrm{Uniform}[25, 31]$ and $\gamma \sim \textrm{Uniform}[6/3, 10/3]$. For Double Pendulum, we chose $m \sim \textrm{Uniform}[0.5, 1.5]$ and $l \sim \textrm{Uniform}[0.5, 1.5]$. For Quadrocopter, we chose independent Gaussians, centered around the ground truth, with a standard deviation of $0.005$. For all experiments, the prior on the observation noise was set to $\sigma \sim \mathrm{InverseGamma}[2, 3]$, except for $\operatorname{SGLD}$ when inferring the Lorenz system. There, we had to fix the noise standard deviation to its ground truth, to get convergence.

For the non-parametric case shown in the main paper, we needed a different strategy, since no ground truth information was available for the weights of the neural dynamics model. Thus, we first trained a deterministic dynamics model. As data, we sampled $100$ tuples $\bm{x}, \dot{\bm{x}}$ equidistantly in time on the trajectory. Note that we implicitly assume access to the ground truth of the dynamics model, i.e. we assume we are provided with accurate, noise free $\dot{\bm{x}}$. The neural dynamics model was then pre-trained on these pairs, until the loss was almost zero (up to $1\mathrm{e}{-5}$). $\operatorname{SGLD}$ and $\operatorname{SGHMC}$ were then provided with Gaussian priors, independent for each component of $\bm{\theta}$, centered around the pre-trained weights, with a standard deviation of 0.1.

\subsection{Number of integrations for prediction}
\label{subsection: number of integration for prediction} $\operatorname{SGLD}$ and $\operatorname{SGHMC}$ both return samples of the parameters of the dynamics model. To obtain uncertainties in the state space at prediction time, each one of these samples needs to be turned into a sample trajectory, by using numerical integration. To obtain maximum accuracy, we would ideally integrate all parameter samples obtained by the chains. However, due to the computational burden inflicted by numerical integration, this is not feasible. We thus need to find a trade-off between accuracy and computational cost, by randomly subsampling the number of available parameter samples.

\begin{figure}[ht]
    \centering
    \includegraphics[width=1.0\linewidth]{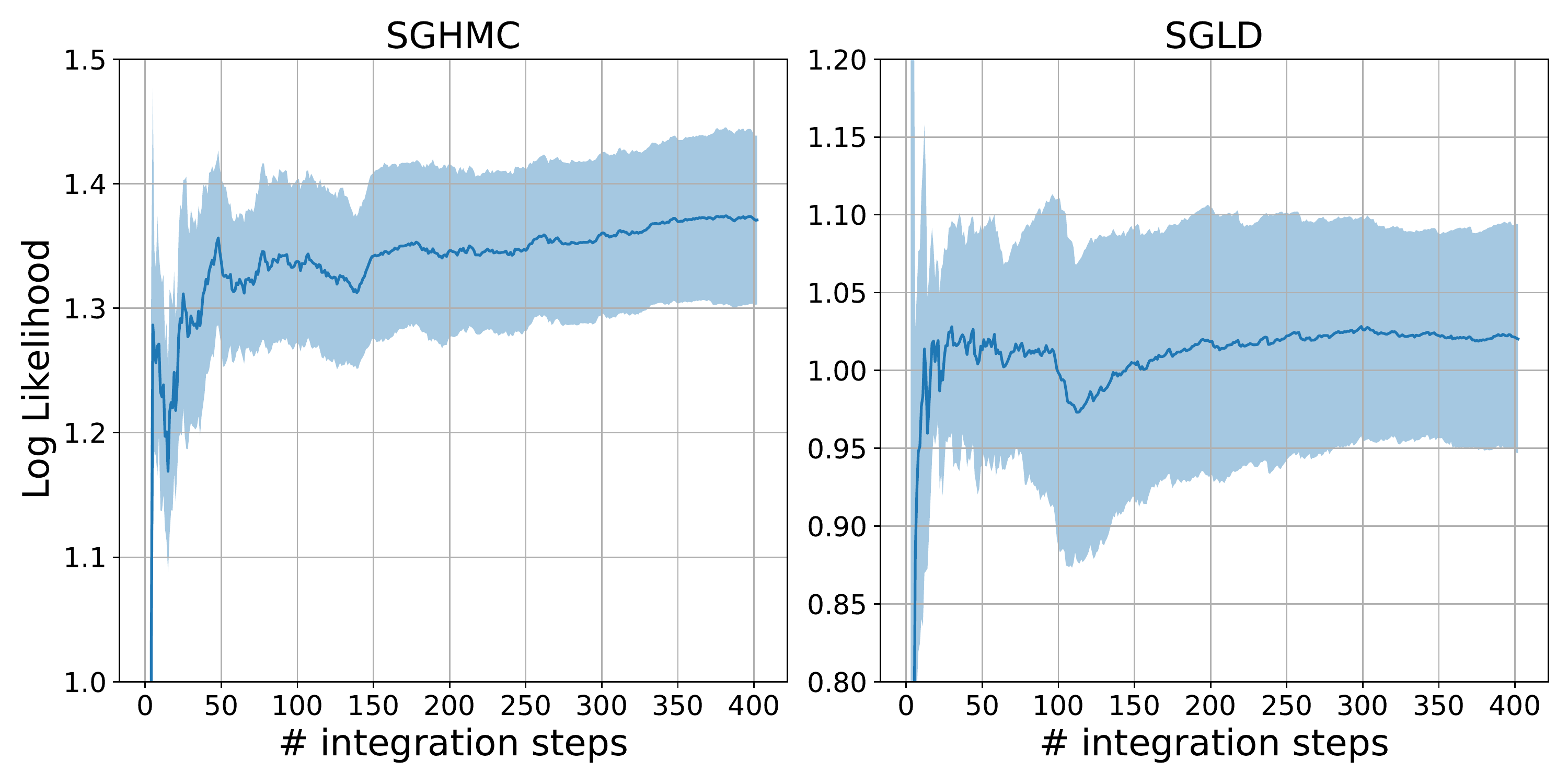}
    \caption{We select the number of sample trajectories for uncertainty prediction to be $200$, since we observe that the log likelihood of the ground truth stops fluctuating  after $200$ steps.}
    \label{figure: number of intergrations}
\end{figure}

In \Cref{figure: number of intergrations} we show how the log likelihood of the ground truth changes with increasing number of sample trajectories on the LV 1 dataset. After initial fluctuations, the log likelihood of the ground truth stabilizes after approximately $200$ steps. To obtain the results of \Cref{table: 1 trajectory training}, we thus chose $200$ integration steps.

\newpage
\section{Scaling to many observations or trajectories}
\label{section: scaling}
Let $N$ be the total number of observations, summed over all training trajectories. In this section, we will analyze the computational complexity of $\operatorname{DGM}$ in terms of $N$ and demonstrate how this can be drastically reduced using standard methods from the literature. For notational compactness, we will assume that the supporting points in $\mathcal{T}$ are at the same locations as the observations in $\mathcal{D}$. However, this is by no means necessary. As long as they are chosen to be constant or stand in a linear relationship to the number of observations, our analysis still holds. We will thus use $\bm{x}$ and $\bm{x}_{\text{supp}}$ and the corresponding quantities interchangeably. Similarly, we will omit the $k$ that was used for indexing the state dimension and assume one-dimensional systems. The extension to multi-dimensional systems is straight forward and comes at the cost of an additional factor $K$.

Fortunately, most components of the loss of $\operatorname{DGM}$ given by \Cref{equation: Simplified Loss} can be calculated in linear time. In particular, it is worth noting that the independence assumption made when calculating the Wasserstein distance in \Cref{equation: Wasserstein approximate} alleviates the need to work with the full covariance matrix and lets us work with its diagonal elements instead. Nevertheless, there are several terms that are not straight forward to calculate. Besides the marginal log likelihood of the observations, these are the posteriors
\begin{align}
    \label{equation: appendix state posterior}
    p_S(\bm{x} | \mathcal{D}, \mathcal{T}) &= \mathcal{N}\left(\bm{x} | \bm{\mu}_{\text{post}}, \bm{\Sigma}_{\text{post}}\right), \\
    \label{equation: appendix deriv posterior}
    p_S(\dot{\mathcal{X}} | \mathcal{D}, \mathcal{T}) &= \normal[\dot{\bm{x}}_{\text{supp}}]{\bm{\mu}_S}{\bm{\Sigma}_S},
\end{align}
where
\begin{align}
    \bm{\mu}_{\text{post}} &= 
    \bm{\mu} + \bm{\mathcal{K}}^T (\bm{\mathcal{K}} + \sigma^2 \bm{I})^{-1}
    (\bm{y} - \bm{\mu}), \\
    \bm{\Sigma}_{\text{post}} &= \bm{\mathcal{K}} - \bm{\mathcal{K}}^T
    (\bm{\mathcal{K}} + \sigma^2 \bm{I})^{-1}\bm{\mathcal{K}},\\
    \bm{\mu}_S &=
    \bm{\dot{\mu}} +  \dot{\bm{\mathcal{K}}}(\bm{\mathcal{K}} + \sigma^2 \bm{I})^{-1}\left(\bm{y} - \bm{\mu}\right), \\
    \bm{\Sigma}_S &= \ddot{\bm{\mathcal{K}}} - \dot{\bm{\mathcal{K}}}(\bm{\mathcal{K}} + \sigma^2\bm I)^{-1}\dot{\bm{\mathcal{K}}}^\top.
\end{align}
Here, \Cref{equation: appendix state posterior} is used for prediction, while its mean is also used in the approximation of \Cref{equation: Wasserstein approximate}. On the other hand, \Cref{equation: appendix deriv posterior} is used directly for \Cref{equation: Wasserstein approximate}. Note that in both cases, we only need the diagonal elements of the covariance matrices, a fact that will become important later on.

In its original form, calculating the matrix inverses of both \Cref{equation: appendix state posterior} and \Cref{equation: appendix deriv posterior} has cubic complexity in $N$. To alleviate this problem, we follow \citet{rahimi2007random} and \citet{angelis2020sleipnir} by using a feature approximation of the kernel matrix and its derivatives. In particular, let $\bm{\Phi} \in \mathbb{R}^{F \times N}$ be a matrix of $F$ random Fourier features as described by \citet{rahimi2007random}. Furthermore, denote $\dot{\bm{\Phi}}$ as its derivative w.r.t. the time input variable, as defined by \citet{angelis2020sleipnir}. We can now approximate the kernel matrix and its derivative versions as
\begin{align}
    \bm{K} \approx \bm{\Phi}^\top \bm{\Phi},
    \quad \dot{\bm{\mathcal{K}}}^\top \approx \dot{\bm{\Phi}}^\top \bm{\Phi},
    \quad \text{and} \quad 
    \ddot{\bm{\mathcal{K}}} \approx \dot{\bm{\Phi}}^\top\dot{\bm{\Phi}}.
\end{align}

Using these approximations, we can leverage the Woodbury idendity to approximate
\begin{equation}
    (\bm{\mathcal{K}} + \sigma^2\bm I)^{-1} \approx
    \frac{1}{\sigma^2} 
    \left[
    \bm{I} - 
    \bm{\Phi}^\top \left(
    \bm{\Phi} \bm{\Phi}^\top + \sigma^2 \bm{I}
    \right)^{-1} \bm{\Phi}
    \right].
\end{equation}
This approximation allows us to invert a $F \times F$ matrix, to replace the inversion of a $N \times N$ matrix. This can be leveraged to calculate
\begin{align}
    \bm{\mu}_S &=
    \bm{\dot{\mu}} +  \dot{\bm{\mathcal{K}}}(\bm{\mathcal{K}} + \sigma^2 \bm{I})^{-1}\left(\bm{y} - \bm{\mu}\right) \\
    &\approx 
    \bm{\dot{\mu}} +
    \frac{1}{\sigma^2} 
    \dot{\bm{\Phi}}^\top\bm{\Phi}
    \left[
    \bm{I} - 
    \bm{\Phi}^\top \left(
    \bm{\Phi} \bm{\Phi}^\top + \sigma^2 \bm{I}
    \right)^{-1} \bm{\Phi}
    \right]
    \left(\bm{y} - \bm{\mu}\right)
    \label{equation: appendix approx mean}
\end{align}

and 
\begin{align}
    \bm{\Sigma}_S &= \ddot{\bm{\mathcal{K}}} - \dot{\bm{\mathcal{K}}}(\bm{\mathcal{K}} + \sigma^2\bm I)^{-1}\dot{\bm{\mathcal{K}}}^\top\\
    & \approx
    \dot{\bm{\Phi}}^\top \dot{\bm{\Phi}} - \frac{1}{\sigma^2} 
    \dot{\bm{\Phi}}^\top\bm{\Phi}
    \left[
    \bm{I} - 
    \bm{\Phi}^\top \left(
    \bm{\Phi} \bm{\Phi}^\top + \sigma^2 \bm{I}
    \right)^{-1} \bm{\Phi}
    \right]
    \bm{\Phi}^\top\dot{\bm{\Phi}}.
    \label{equation: appendix approx cov}
\end{align}

Evaluating the matrix multiplications of \Cref{equation: appendix approx mean} in the right order leads to a computational complexity of $\mathcal{O}(NF^2 + F^3)$. Similarly, the diagonal elements of the covariance given by \Cref{equation: appendix deriv posterior} can be calculated with the same complexity, by carefully summarizing everything in between $\dot{\bm{\Phi}}^\top$ and $\dot{\bm{\Phi}}$ as one $F \times F$ matrix and then calculating the $N$ products independently.

Since the components of \Cref{equation: appendix state posterior} have the exact same form as the components of \Cref{equation: appendix deriv posterior}, they can be approximated in the exact same way to obtain the exact same computational complexity. Thus, the only components that need further analysis are the components of the marginal log likelihood of the observations, particularly 
\begin{equation}
    \bm{y}^\top (\bm{\mathcal{K}} + \sigma^2\bm I)^{-1} \bm{y} \approx 
    \bm{y}^\top \frac{1}{\sigma^2} 
    \left[
    \bm{I} - 
    \bm{\Phi}^\top \left(
    \bm{\Phi} \bm{\Phi}^\top + \sigma^2 \bm{I}
    \right)^{-1} \bm{\Phi}
    \right] \bm{y}
\end{equation}
and 
\begin{align}
    \text{logdet}(\bm{\mathcal{K}} + \sigma^2\bm I) &\approx 
    \text{logdet}(\bm{\Phi}^\top \bm{\Phi} + \sigma^2\bm{I}) \\
    & \approx
    \text{logdet}(\bm{\Phi} \bm{\Phi}^\top + \sigma^2\bm{I}) + (N-F)\text{log}(\sigma^2).
\end{align}
In the last line, we used the fact that the nonzero eigenvalues of the transposed of a matrix stay the same.

Combining all these tricks, it is clear that the overall complexity of $\operatorname{DGM}$ can be reduced to $\mathcal{O}(NF^2 + F^3)$. Since $F$ is a constant controlling the quality of the approximation scheme and is usually chosen to be constant, we thus get essentially linear computational complexity in the number of observations. Note that these derivations are completely independent of what scheme is chosen to obtain the feature matrix $\bm{\Phi}$. For ease of implementation, we opted for random Fourier features though in our experiments.

\paragraph{Experimental Proof of Concept} To demonstrate that this approximation scheme can be used in the context of $\operatorname{DGM}$, we tested it on the multi-trajectory experiment of Lotka Volterra. To this end, we increased the grid from $10$ points per dimension to $25$, leading to a total number of $3125$ observations instead of $500$. As an approximation, we used $50$ random Fourier features. Through this approximation, $\operatorname{DGM}$ became slightly more sensitive to the optimization hyperparameters. Nevertheless, it reached comparable accuracy within roughly $440$ seconds of training, compared to the $408$ seconds needed to train the approximation free version on LV 100.
\newpage
\section{Additional experiments}
\label{section: additional experiments}

In this section, we first show the state predictions of $\operatorname{DGM}$ on the datasets with multiple trajectories. Then, we compare $\operatorname{DGM}$ with $\operatorname{SGLD}$ and $\operatorname{SGHMC}$ for the parametric case, i.e. where we assume to have access to the true parametric form of the dynamics. Since most datasets have too many trajectories to be fully visualized, we show a random subset instead.

\subsection{Sample plots from trained trajectories}
\label{subsection: sample plots from trained trajectories}

\begin{figure}[H]
    \centering
    \includegraphics[width=0.78\linewidth]{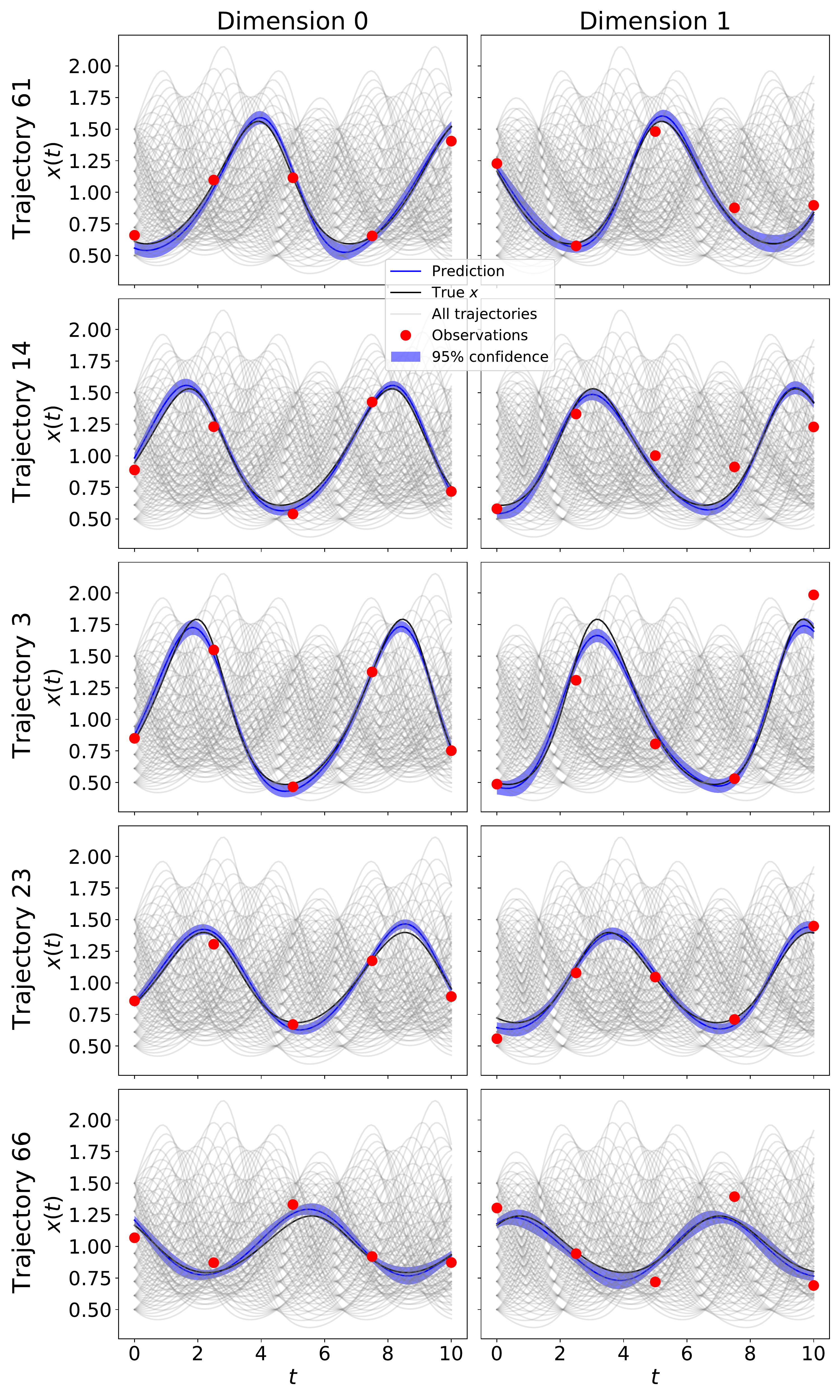}
    \caption{$\operatorname{DGM}$'s prediction on $5$ randomly sampled training trajectories of LV 100.}
\end{figure}

\begin{figure}[H]
    \centering
    \includegraphics[width=0.9\linewidth]{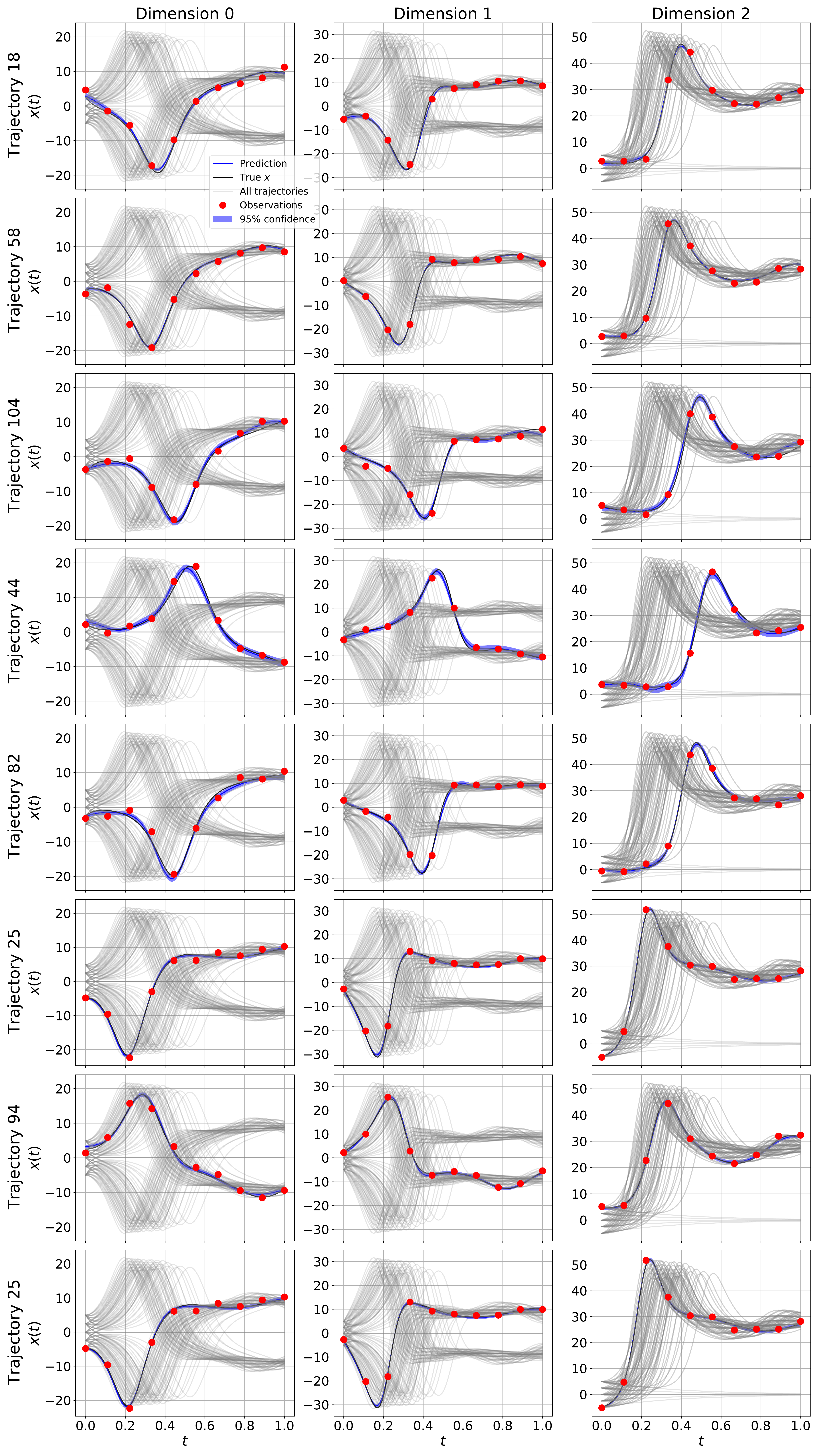}
    \caption{$\operatorname{DGM}$'s prediction on $8$ randomly sampled training trajectories of LO 125.}
\end{figure}

\begin{figure}[H]
    \centering
    \includegraphics[width=0.9\linewidth]{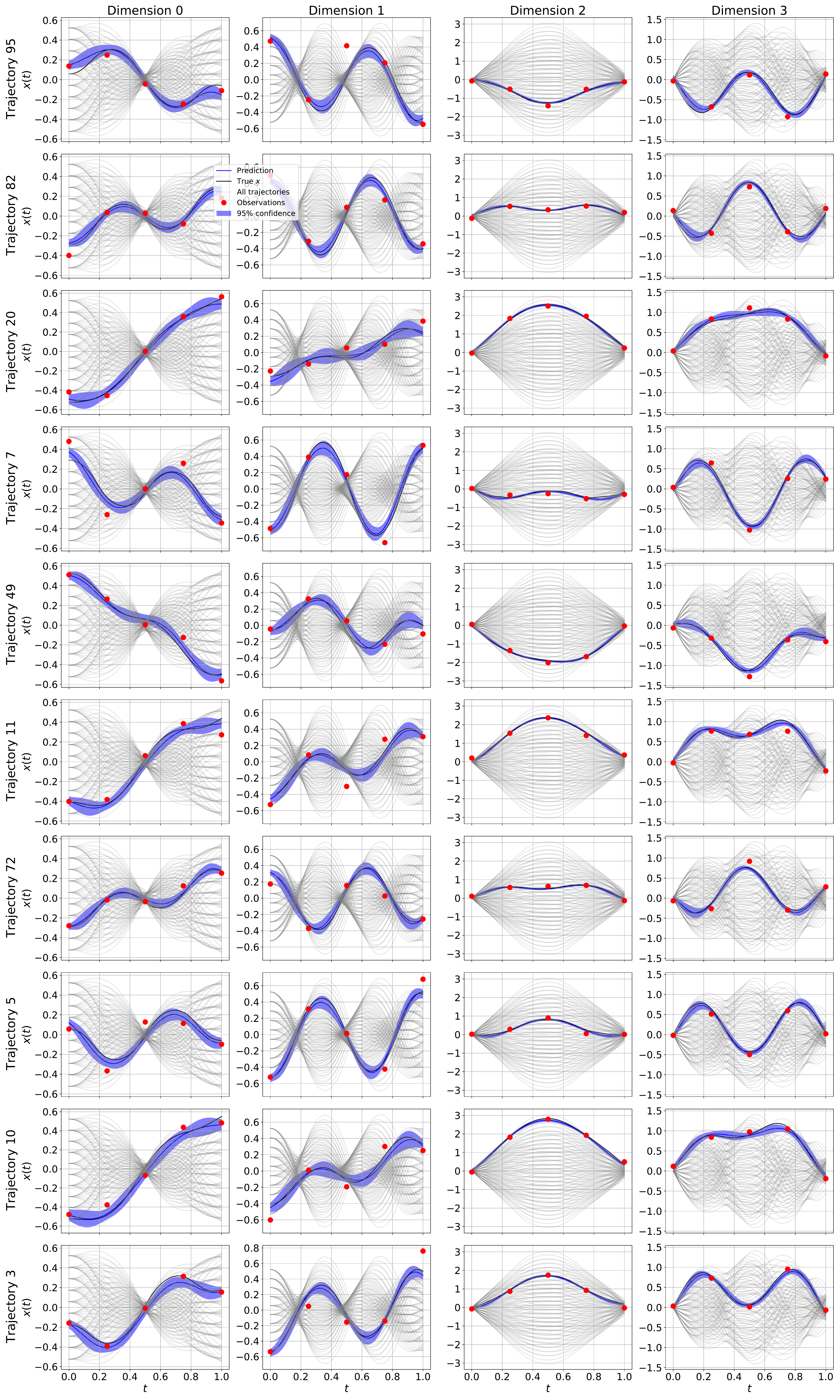}
    \caption{$\operatorname{DGM}$'s prediction on $10$ randomly sampled training trajectories of DP 100.}
\end{figure}

\begin{figure}[H]
    \centering
    \includegraphics[width=0.7\linewidth]{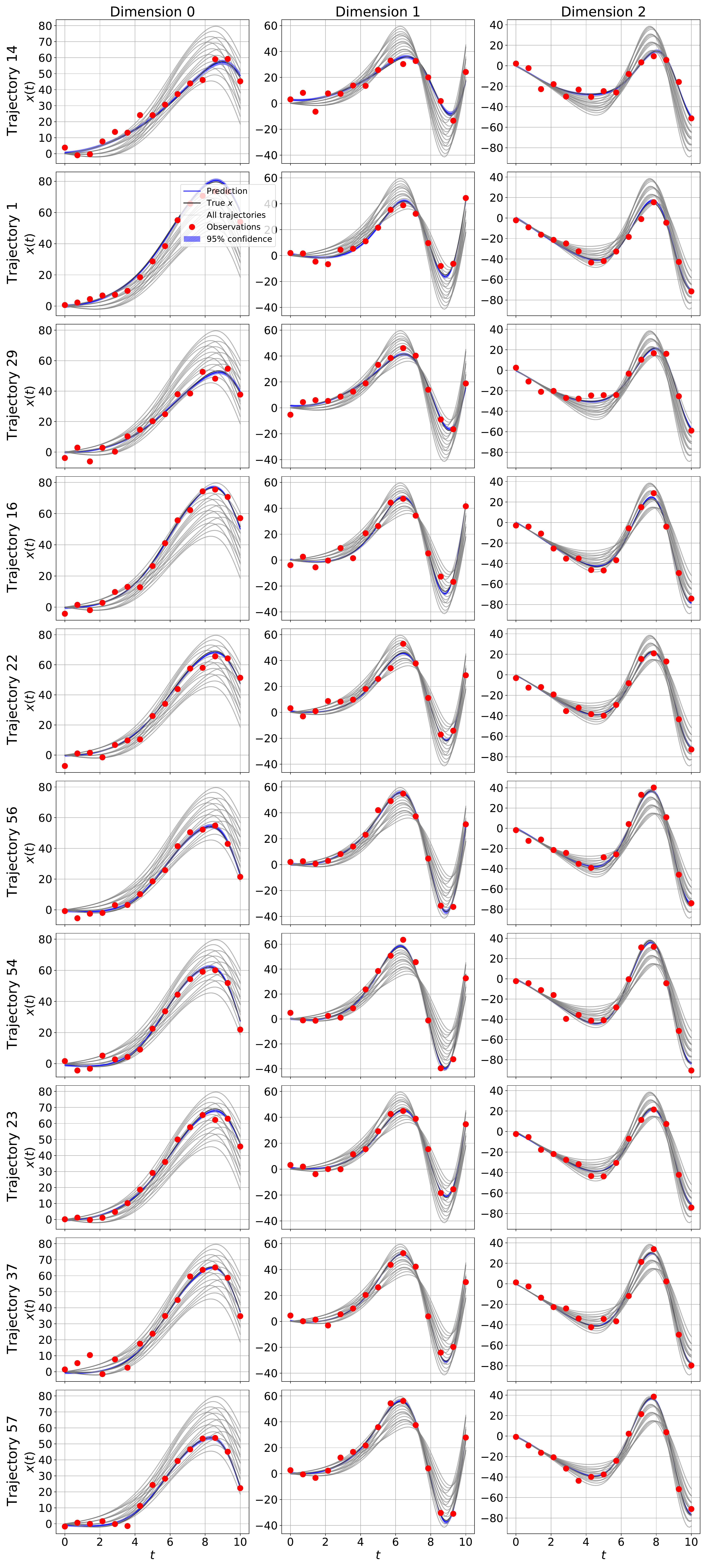}
    \caption{$\operatorname{DGM}$'s prediction on $10$ randomly sampled training trajectories of QU 64, for state dimensions $0$-$2$.}
\end{figure}

\begin{figure}[H]
    \centering
    \includegraphics[width=0.7\linewidth]{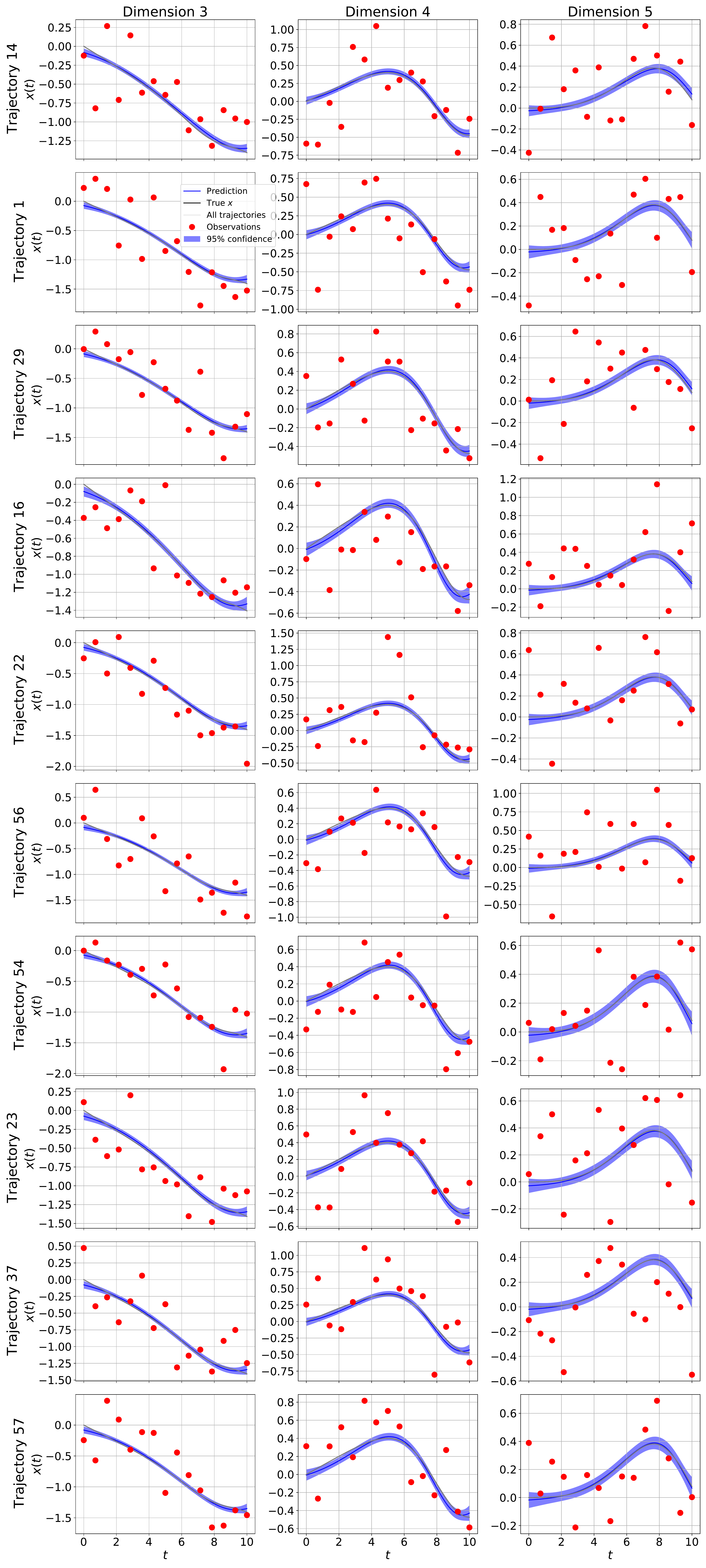}
    \caption{$\operatorname{DGM}$'s prediction on $10$ randomly sampled training trajectories of QU 64, for state dimensions $3$-$5$.}
\end{figure}

\begin{figure}[H]
    \centering
    \includegraphics[width=0.7\linewidth]{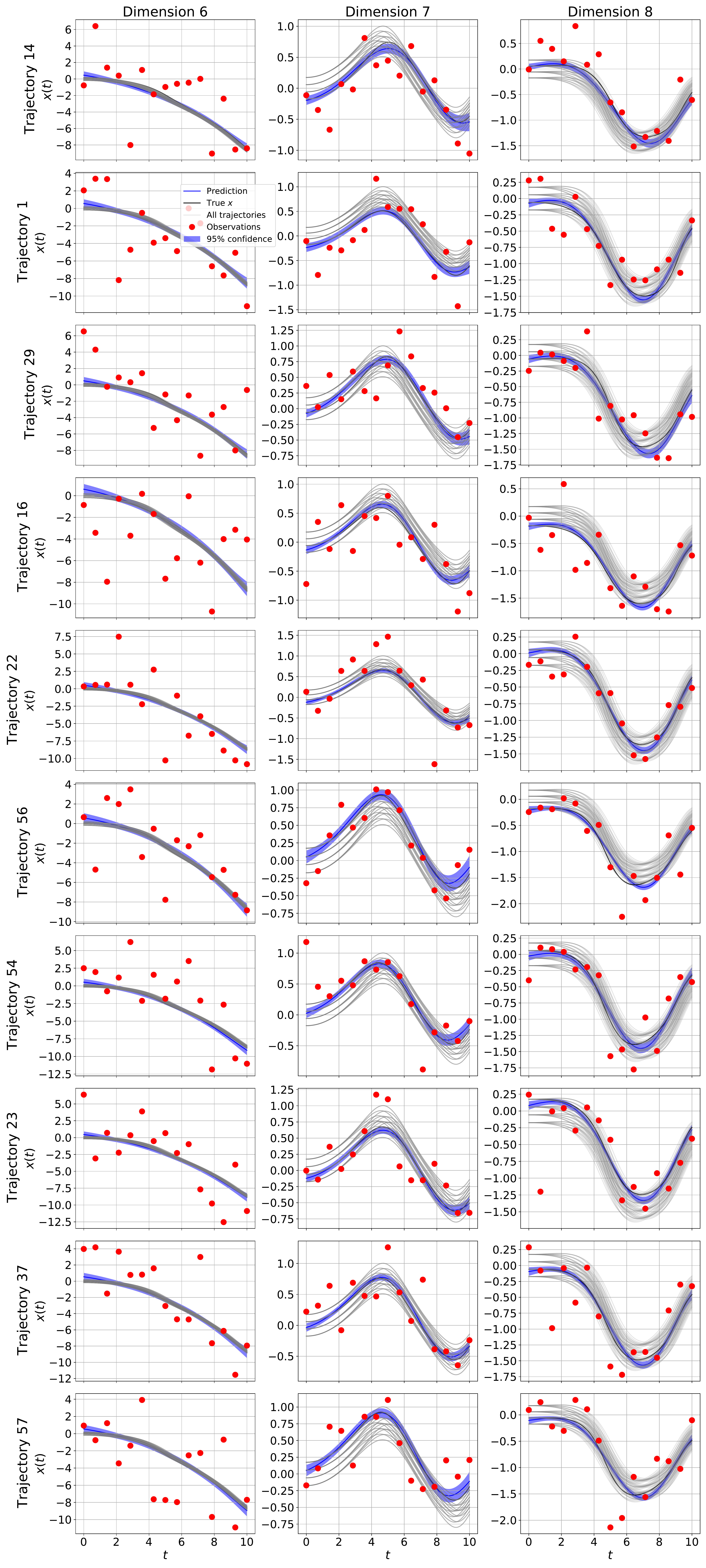}
    \caption{$\operatorname{DGM}$'s prediction on $10$ randomly sampled training trajectories of QU 64, for state dimensions $6$-$8$.}
\end{figure}

\begin{figure}[H]
    \centering
    \includegraphics[width=0.7\linewidth]{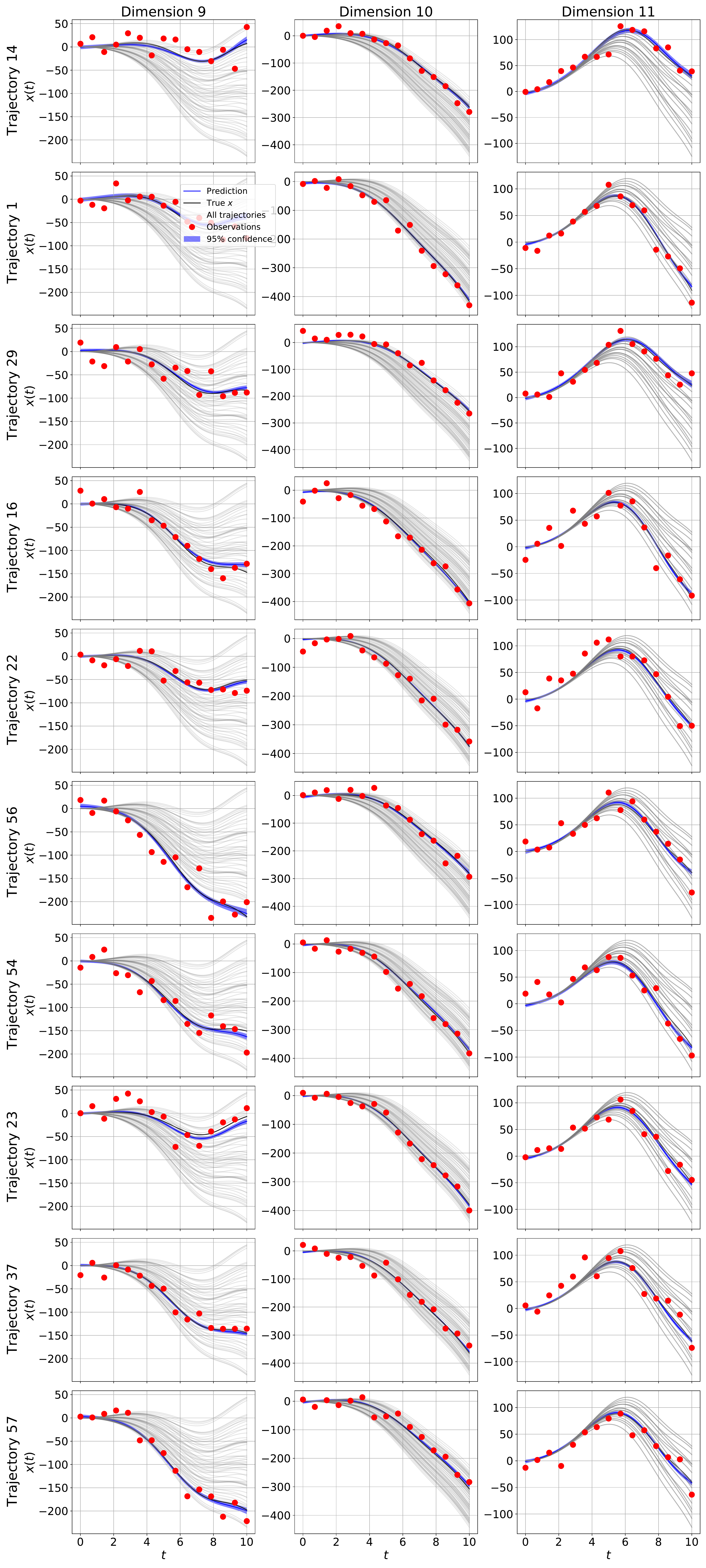}
    \caption{$\operatorname{DGM}$'s prediction on $10$ randomly sampled training trajectories of QU 64, for state dimensions $9$-$11$.}
\end{figure}

\subsection{Sample plots from test trajectories}
\label{subsection: sample plots from test trajectories}

\begin{figure}[H]
    \centering
    \includegraphics[width=0.4\linewidth]{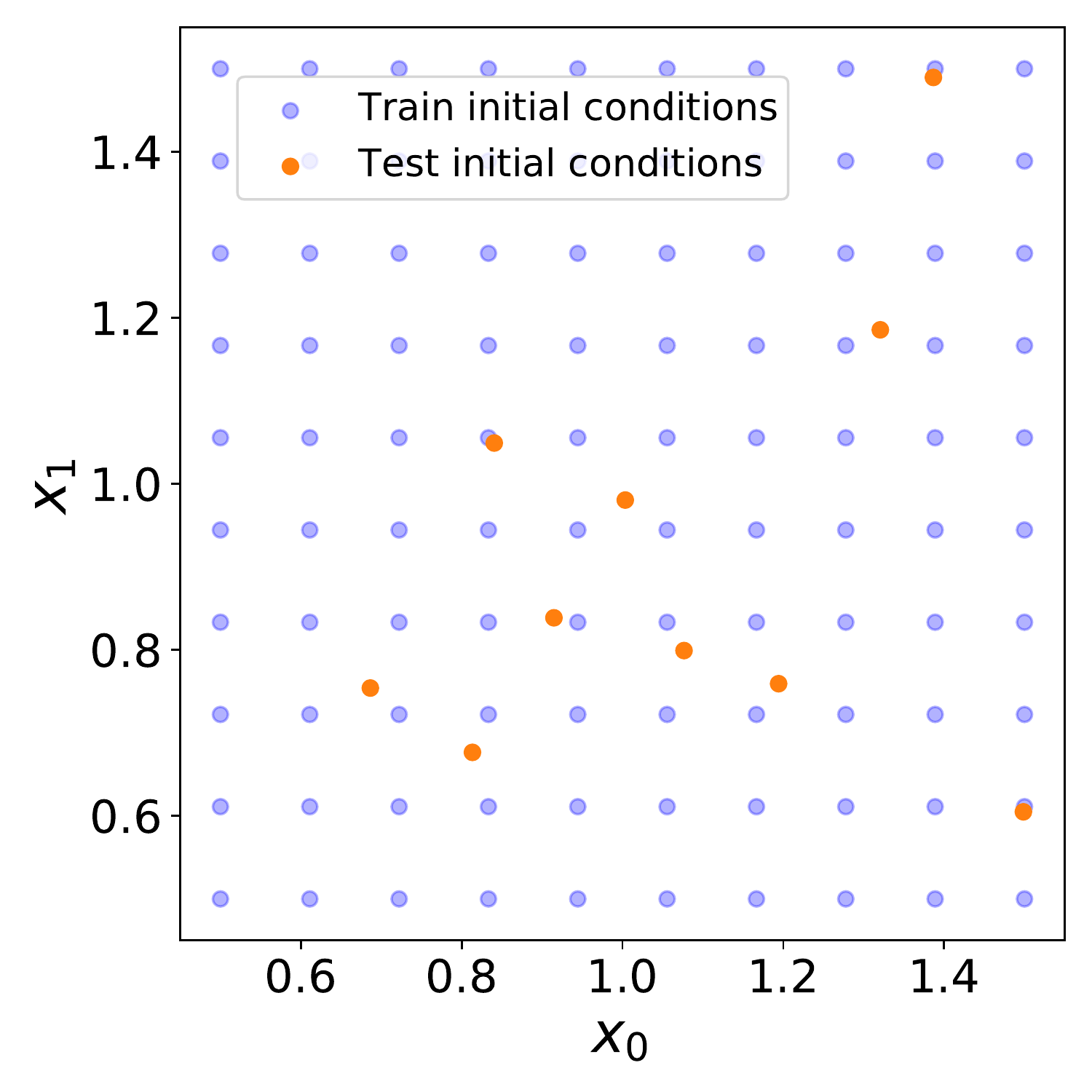}
    \caption{\looseness-1 Placement of the initial conditions for the train and test trajectories of the LV 100 dataset. We selected the initial conditions for the train trajectories by gridding $[0.5, 1.5]^2$ with $10$ points in every dimension. We select initial conditions for test trajectories independently, uniformly at random from the cube $[0.5, 1.5]^2$.}
    \label{figure: placement of initial conditions }
\end{figure}

Here, we show $\operatorname{DGM}$'s predictions on the test trajectories used to test generalization, as introduced in \Cref{section: dataset description}. Since LV 100 is a two dimensional system, we also show the placement of the train and test initial conditions in \Cref{figure: placement of initial conditions }.

\begin{figure}[H]
\centering
\vspace{-1cm}
\begin{subfigure}{.5\textwidth}
  \centering
  \includegraphics[width=\linewidth]{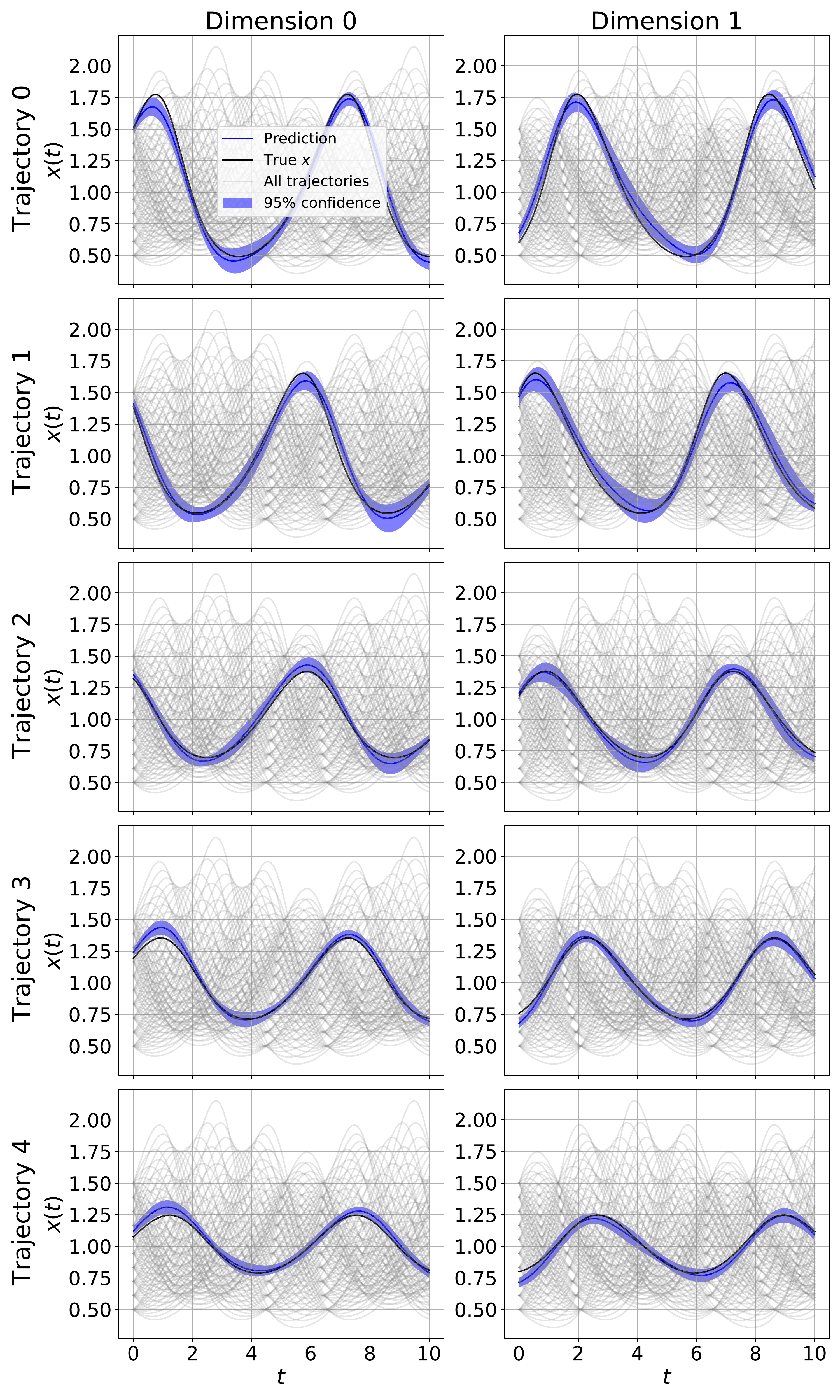}
\end{subfigure}%
\begin{subfigure}{.5\textwidth}
  \centering
  \includegraphics[width=\linewidth]{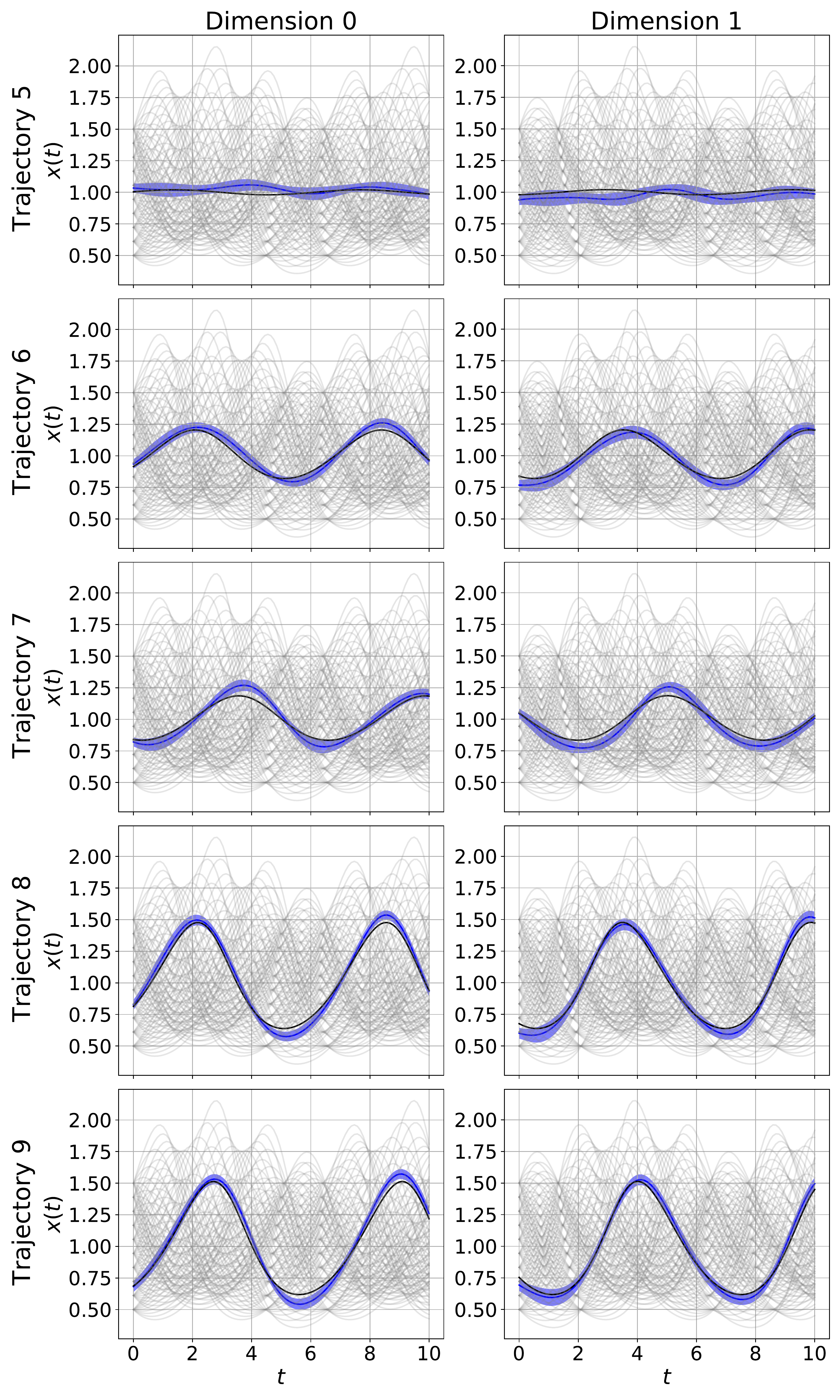}
\end{subfigure}
\caption{$\operatorname{DGM}$'s prediction on $10$ randomly sampled test trajectories for the LV 100 dataset.}
\end{figure}

\begin{figure}[H]
    \centering
    \includegraphics[width=0.7\linewidth]{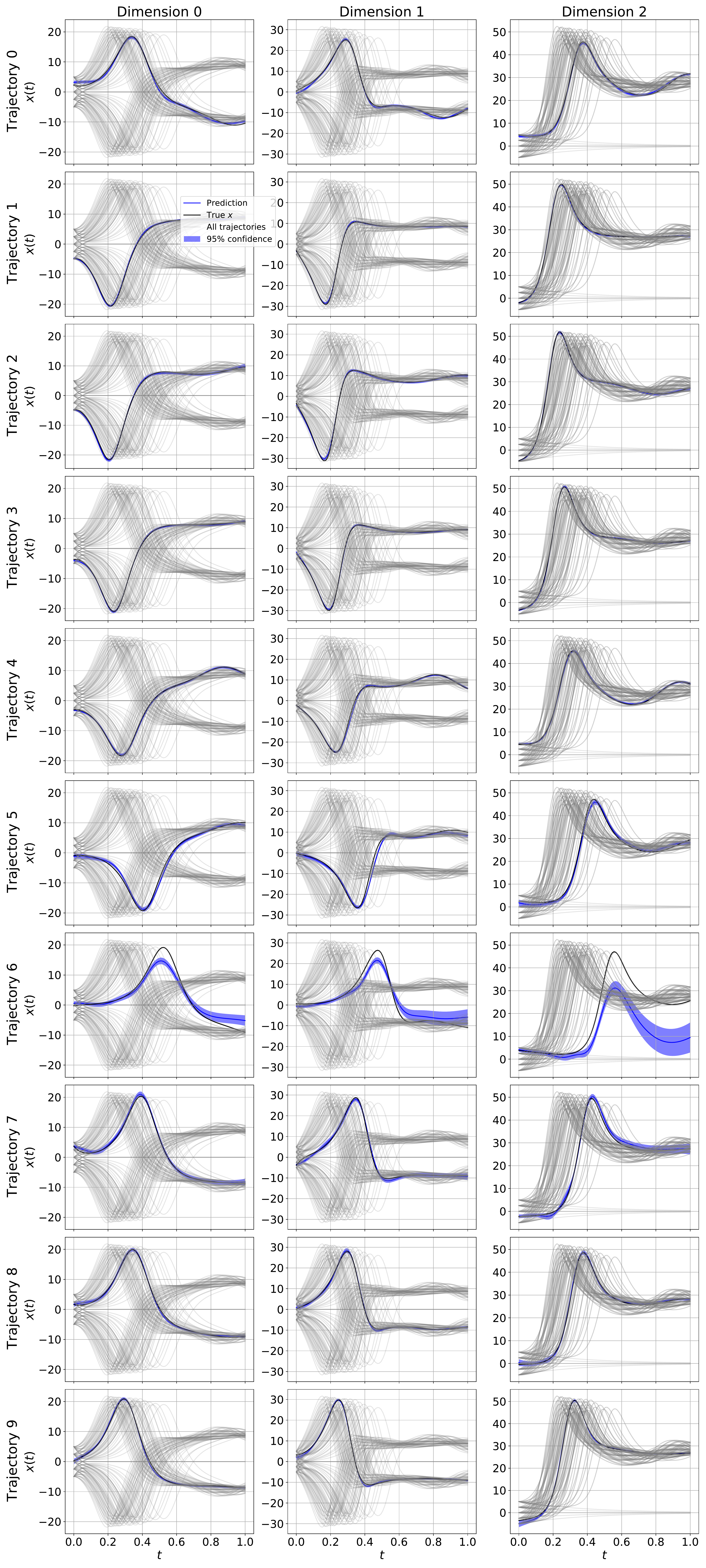}
    \caption{$\operatorname{DGM}$'s prediction on $10$ randomly sampled test trajectories for the LO 125 dataset.}
\end{figure}

\begin{figure}[H]
    \centering
    \includegraphics[width=0.9\linewidth]{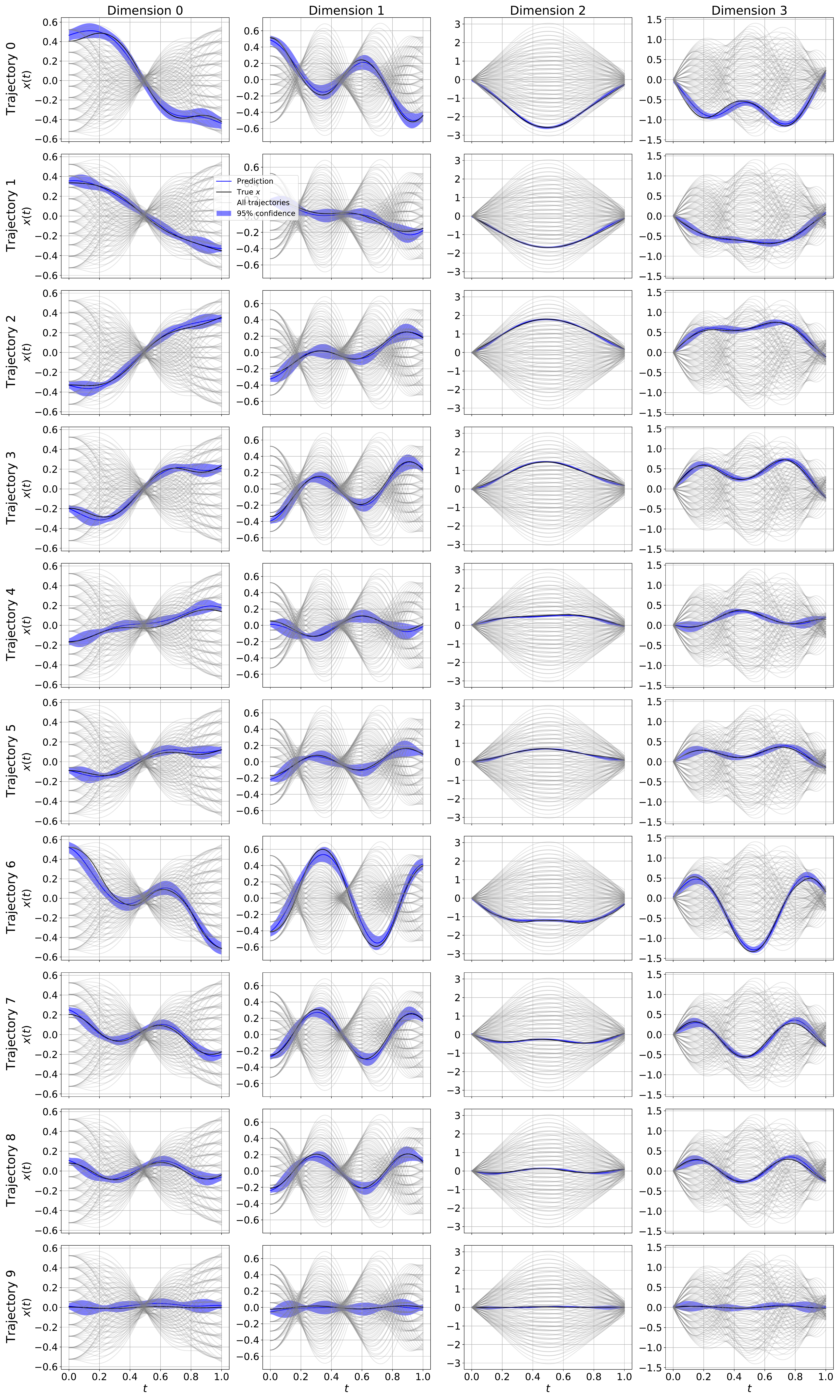}
    \caption{$\operatorname{DGM}$'s prediction on $10$ randomly sampled test trajectories for the DP 100 dataset.}
\end{figure}

\begin{figure}[H]
    \centering
    \includegraphics[width=0.7\linewidth]{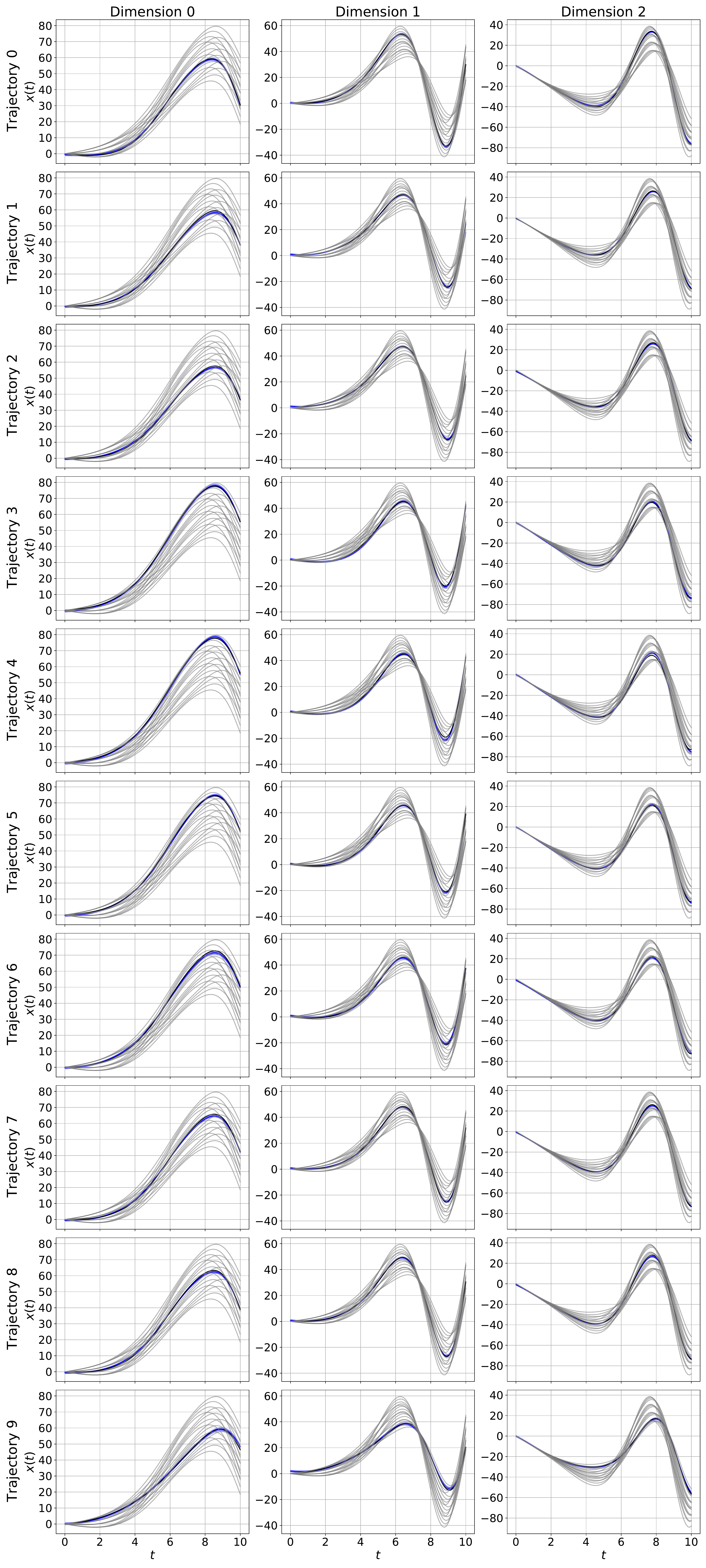}
    \caption{$\operatorname{DGM}$'s prediction on $10$ randomly sampled test trajectories of QU 64, for state dimensions $0$-$2$.}
\end{figure}

\begin{figure}[H]
    \centering
    \includegraphics[width=0.7\linewidth]{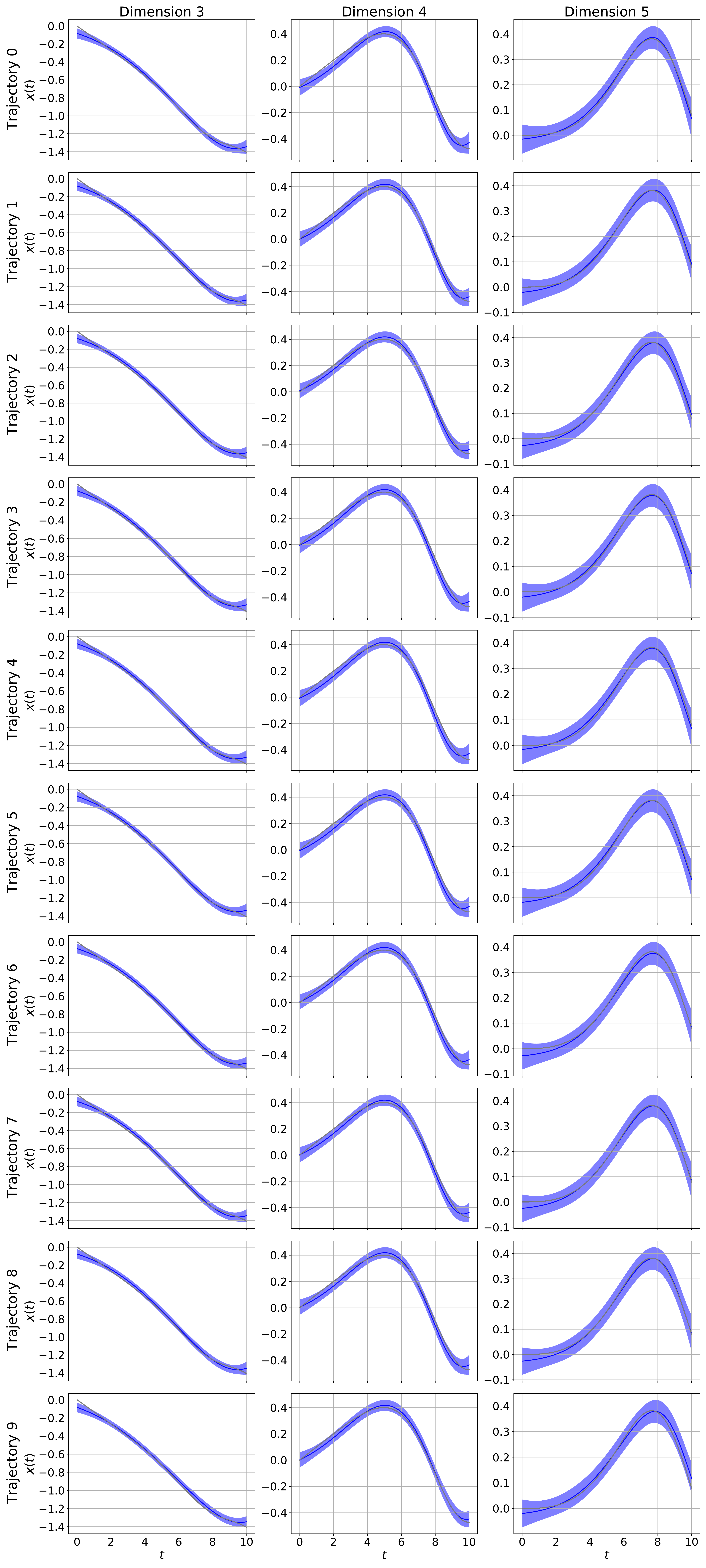}
    \caption{$\operatorname{DGM}$'s prediction on $10$ randomly sampled test trajectories of QU 64, for state dimensions $3$-$5$.}
\end{figure}

\begin{figure}[H]
    \centering
    \includegraphics[width=0.7\linewidth]{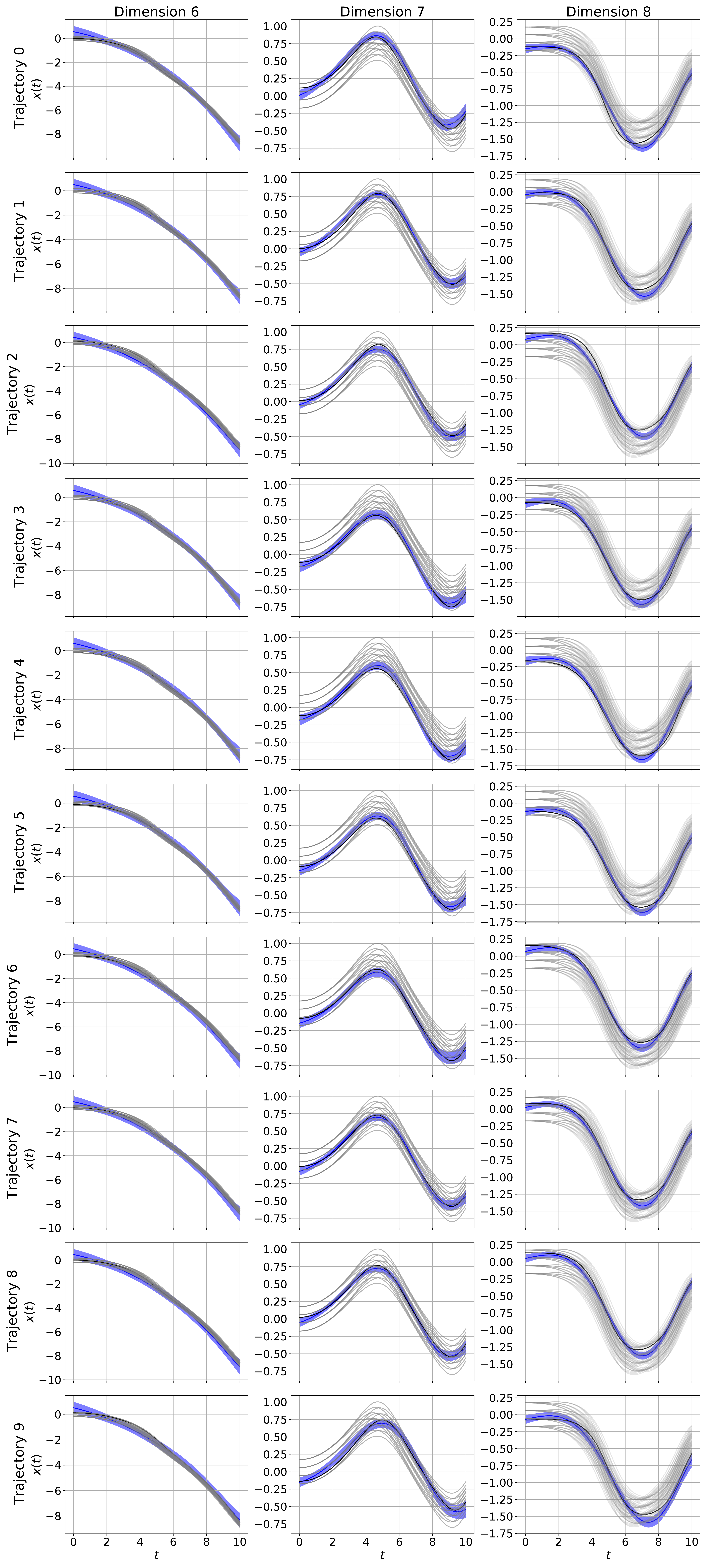}
    \caption{$\operatorname{DGM}$'s prediction on $10$ randomly sampled test trajectories of QU 64, for state dimensions $6$-$8$.}
\end{figure}

\begin{figure}[H]
    \centering
    \includegraphics[width=0.7\linewidth]{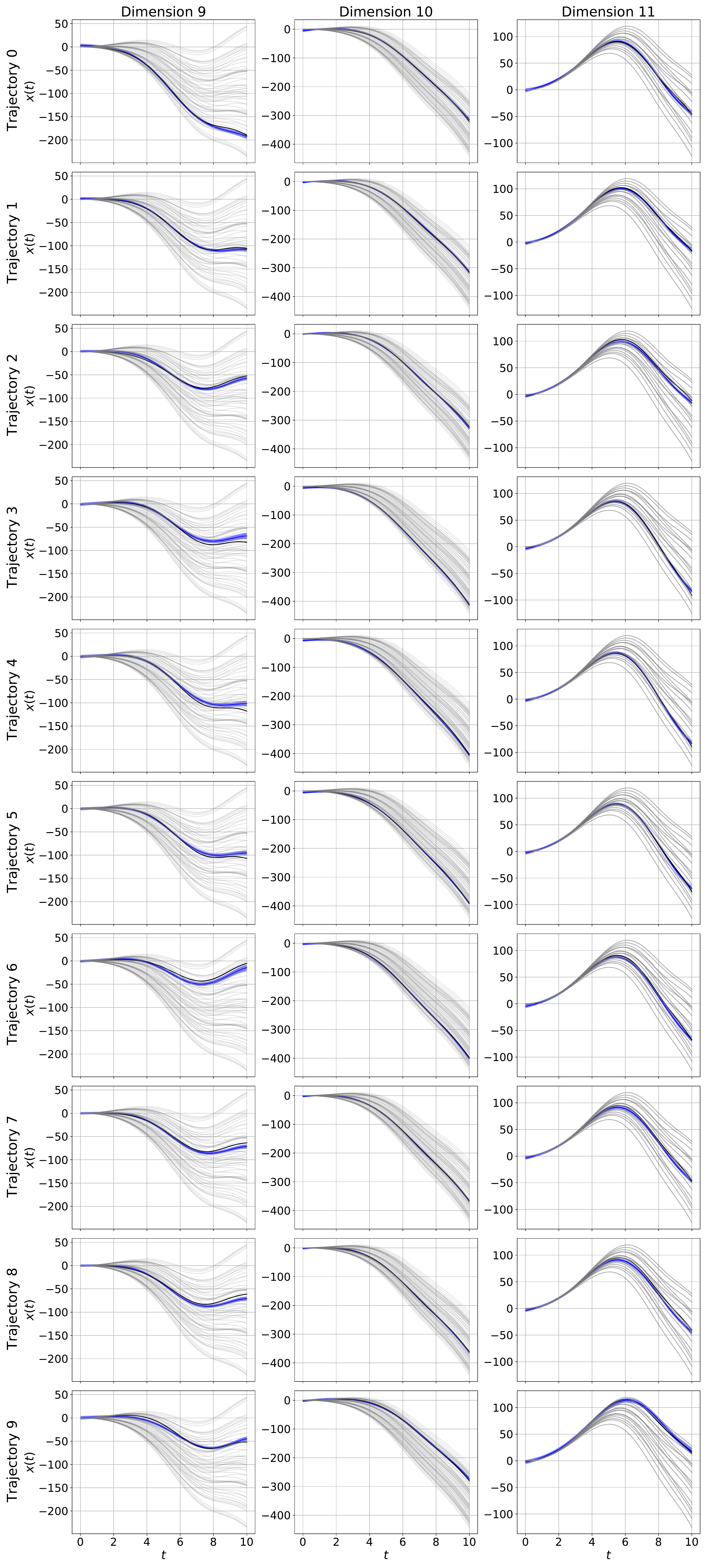}
    \caption{$\operatorname{DGM}$'s prediction on $10$ randomly sampled test trajectories of QU 64, for state dimensions $9$-$11$.}
\end{figure}

\subsection{Comparison with parameteric integration}
\label{subsection: comparison with parameteric integration}

In this subsection, we compare $\operatorname{DGM}$ against $\operatorname{SGLD}$ and $\operatorname{SGHMC}$ in the parametric setting, i.e. where we assume access to the parametric form of the true dynamics $\bm{f}(\bm{x}, \bm{\theta})$. Despite serious tuning efforts outlined in \Cref{subsection: hyperparameters selection}, we were unable to make $\operatorname{SGLD}$ and $\operatorname{SGHMC}$ perform on any multitrajectory experiments except for Lotka Volterra 100. As can be seen in \Cref{Table: parametric comparisons log likelihood}, the sampling based methods seem to perform quite well. However, it should be noted that we were unable to get a stable performance without using ground truth information, as outlined in \Cref{subsection: hyperparameters selection}. Given this caveat and the results of the non-parametric case in the main paper, we conclude the following. If strong and accurate expert knowledge is available that can be used to fix strong priors on simple systems, the sampling-based approaches are certainly a good choice. For more complex systems or in the absence of any expert knowledge, $\operatorname{DGM}$ seems to have a clear edge.

\begin{table}[H]
\centering
\caption{Log likelihood of the ground truth of $100$ points on the training trajectories. $\operatorname{SGHMC}$ and $\operatorname{SGLD}$ were provided with strong, ground-truth-inspired priors and received an extensive hyperparameter sweep using the ground truth as metric. Nevertheless, $\operatorname{DGM}$ performs decently in comparison, without using neither priors nor ground truth.}
\label{Table: parametric comparisons log likelihood}
\begin{tabular}{@{}lccc@{}}
\toprule
        & \multicolumn{3}{c}{Log Likelihood}           \\ \cmidrule{2-4}
            & $\operatorname{DGM}$ & $\operatorname{SGLD}$ & $\operatorname{SGHMC}$ \\
LV 1   & $1.98\pm 0.18$       & $\mathbf{3.07 \pm 0.685}$      & $3.06 \pm 0.517$ \\
LO 1           & $-0.52 \pm 0.09$     & $\mathbf{2.01 \pm 0.548}$      & F                \\
DP 1  & $2.16 \pm 0.13$      & $\mathbf{3.43 \pm 0.396}$      & $2.96 \pm 0.795$ \\
QU 1     & $0.71 \pm 0.07$      & $\mathbf{2.42 \pm 0.322}$      & $1.38 \pm 0.00 $ \\
LV 100 & $1.85 \pm 0.11$      & $\mathbf{4.28 \pm 0.184}$      & $4.26 \pm 0.178$ \\ \bottomrule
\end{tabular}
\end{table}

\end{document}